\documentclass{article}

\usepackage[preprint]{neurips_2026}

\usepackage[utf8]{inputenc} 
\usepackage[T1]{fontenc}    
\usepackage{hyperref}       
\usepackage{url}            
\usepackage{booktabs}       
\usepackage{amsfonts}       
\usepackage{nicefrac}       
\usepackage{microtype}      
\usepackage{xcolor}         

\usepackage{graphicx}
\usepackage{subcaption}
\usepackage{mathtools}
\usepackage{amsthm}
\usepackage{listings}
\usepackage{xr}
\usepackage{amsmath}
\usepackage{physics}
\usepackage{multirow}
\usepackage{colortbl}
\usepackage{rotating}
\usepackage{enumitem}
\usepackage{lipsum}
\usepackage{xspace}
\usepackage{algorithm}
\usepackage{algpseudocode}
\usepackage{amssymb}
\usepackage{pifont}
\newcommand{\cmark}{\ding{51}}
\newcommand{\xmark}{\ding{55}}
\usepackage{makecell}

\usepackage{tcolorbox}
\tcbuselibrary{skins, listings, breakable}
\usepackage[scaled=0.9]{beramono} 
\newcommand{\myconsolas}{\ttfamily}

\newcommand{\papername}{\textbf{RoboWits}\xspace}

\usepackage{wrapfig}
\usepackage{titletoc}  

\definecolor{codecolor}{HTML}{B9C78D}
\colorlet{xx}{codecolor}

\newtcblisting{codebox}[2]{
  colback=codecolor!5,
  colframe=codecolor!80!black,
  listing only, 
  listing options={ 
    numbers=left,
    basicstyle=\small\myconsolas,
    breaklines=true,
    numbersep=2mm,
  },
  left=5mm,
  enhanced,
  title=#2, 
  breakable
}

\newtcblisting{minicodebox}[2]{
  colback=codecolor!5,
  colframe=codecolor!80!black,
  listing only, 
  listing options={ 
    basicstyle=\tiny\myconsolas,
    breaklines=true,
  },
  left=0mm,
  right=0mm,
  top=0mm,
  bottom=0mm,
  enhanced,
  title=#2
}

\title{RoboWits: Unexpected Challenges for \\ Robotic Creative Problem Solving}

\author{%
  Chunru Lin$^{1}$\thanks{denotes equal contribution.}, Hongxin Zhang$^{1*}$, Fenghao Yu$^{1}$, Zhehuan Chen$^{1}$\\
  \textbf{Thomas L. Griffiths$^{2}$, Yejin Choi$^{3}$, David Held$^{4}$, Chuang Gan$^{1}$} \\
$^1$ University of Massachusetts Amherst $^2$ Princeton University\\
$^3$ Stanford University $^4$ Carnegie Mellon University \\
  \texttt{\{chunrulin,hongxinzhang,chuangg\}@umass.edu} \\
}

\begin{document}

\maketitle

\begin{abstract}
  The ability to reason, adapt, and creatively solve problems under unexpected challenges is essential for robots operating in real-world environments. However, current robotic benchmarks primarily emphasize skill-level execution and provide limited insight into such cognitive reasoning capabilities. We introduce \papername, a bi-manual robotic benchmark designed to systematically evaluate cognitive reasoning, creative tool use, and robustness to unexpected conditions. To enable scalable construction of high-quality reasoning-centric unexpected scenarios, we propose an automated task generation pipeline formulated as a multi-agent cooperative framework, comprising agents for seed task generation and verification, metric generation, scene generation, and task mutation. Using the pipeline, we curated 30 diverse seed tasks and 200 tasks with mutations and graded difficulty across geometry, material, and assembly-based reasoning. We benchmark popular robot policies, pre-trained VLAs, a zero-shot policy built on a frontier VLM, and oracle-state planners.
  Our results reveal a significant performance gap: while pre-trained VLAs exhibit preliminary success on seed tasks after fine-tuning, they struggle to perform on mutated tasks, implying their brittleness in manipulation tasks requiring reasoning, strategy adaptation, and robustness to deceptive or constrained environments. A zero-shot policy built on \texttt{GPT-6 Astra} outperforms all fine-tuned VLAs, yet still loses about a fifth of its success on mutated tasks, showing that stronger foundation models alone do not close the robustness gap. See our project page\footnote{\url{https://umass-embodied-agi.github.io/RoboWits/}} for videos.
\end{abstract}

\section{Introduction}
\label{sec:intro}

The ability to creatively solve problems and adapt to unexpected challenges is fundamental to intelligent behavior. In everyday environments, humans and animals routinely encounter situations where initial strategies fail, tools behave differently than expected, or misleading alternatives are present. Success in such settings relies not only on executing motor skills, but also on reasoning about object geometry, material properties, and physical constraints, and flexibly revising plans when assumptions are violated. For robots to operate reliably in the real world, they must demonstrate similar cognitive problem-solving capabilities.

\begin{figure}[tb]
    \centering
    \includegraphics[width=\linewidth]{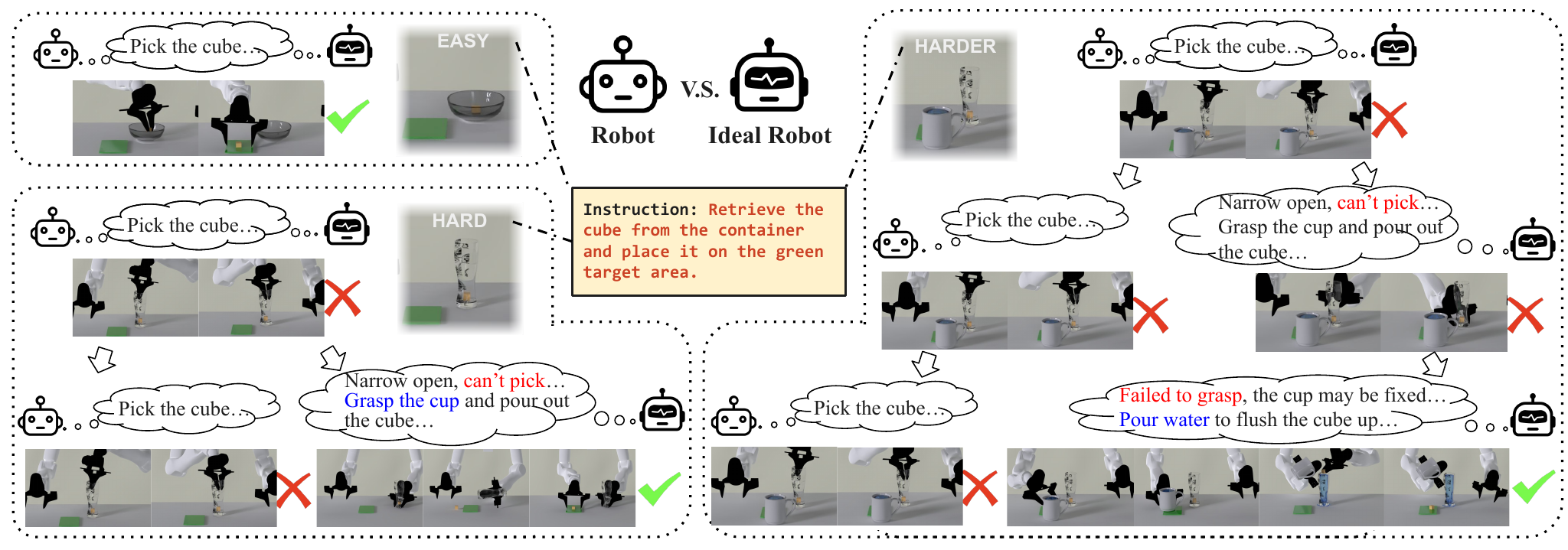}
    \caption{\textbf{Creative problem solving under unexpected challenges}. The figure contrasts the reasoning and execution processes of a standard robot (left) and an ideal robot (right) performing tasks under the same instruction but escalating difficulty. While both succeed in an \textit{Easy} setting (top-left), the standard robot gets stuck repeating direct but unfeasible actions when the cube is trapped deep inside a cup (\textit{Hard}, bottom-left), or when the cup is firmly affixed to the table (\textit{Harder}, right). In contrast, an ideal robot demonstrates true creative problem-solving by actively reasoning through unexpected failures to dynamically discover and execute novel recovery strategies.}
    \label{fig:teaser}
\end{figure}

Despite significant progress in robotic manipulation, current robotic benchmarks largely focus on skill-level execution under well-specified conditions~\cite{james2020rlbench,liu2023libero,geng2025roboverse,chen2025robotwin}, such as grasping, pushing, or predefined tool use~\cite{xu2023creative}. As shown in Figure~\ref{fig:teaser}, VLAs mastering skill-level execution only struggle with small changes in the scene, while VLAs thinking out of the box could reason about the constraints implied in the scene and creatively adapt strategies. While recent VLAs showcase impressive tool-use and failure-recovery demonstrations, such as cardboard box assembly~\cite{intelligence2025pi06vlalearnsexperience} and small-object collection with a paper plate~\cite{generalist2025gen0}, there is no unified benchmark that systematically evaluates a robot’s ability to reason, adapt, and handle unexpected challenges. As a result, the reasoning capabilities of current VLA models are not well understood, and the absence of systematic evaluation makes it difficult to diagnose their failure modes or identify the specific reasoning components that must be improved to advance toward more capable, reasoning-driven VLAs.

To address this gap, we introduce \papername, a new bi-manual robotic benchmark that explicitly targets evaluating cognitive reasoning, creative tool use, and robustness to unexpected challenges. The benchmark consists of manipulation tasks that require geometric, material, and assembly-based reasoning, where naïve solutions are often invalidated or inefficient. Tasks are organized across multiple difficulty levels, enabling fine-grained evaluation of how robotic models scale with increasing reasoning complexity.

Designing such tasks manually is time-consuming and extremely challenging for even human designers. To produce diverse, high-quality tasks at scale, we propose an automated task generation pipeline formulated as a multi-agent cooperative framework. The pipeline decomposes task construction into specialized agents with different roles: a \textit{seed task generator} that proposes cognitively challenging task specifications, a \textit{task mutation generator} that expands each seed into diverse task mutations where original solution are blocked by small scene changes, a \textit{task metric generator} that produces executable evaluation criteria, a \textit{task verifier} that ensures feasibility, simulatability, and necessity of reasoning, and a \textit{scene generator} that constructs realistic and physically consistent environments. Through structured cooperation among these agents, the pipeline automatically produces a diverse and scalable set of cognitively challenging tasks that would be difficult to design by hand. We construct 30 seed tasks spanning geometry, material, and assembly-based reasoning, and expand them into 200 tasks with graded difficulty and full evaluation code. We also collect 50 human-teleoperated demonstrations for 10 of the seed tasks to facilitate benchmarking.

While cognitive reasoning is arguably learnable by imitation, existing robotic policy models remain unable to reliably perform unseen tasks in a zero-shot manner, even after large-scale pre-training. We evaluate a representative suite of models on \papername, including imitation learning baselines, pre-trained Vision–Language–Action (VLA) models, a zero-shot policy built on the Vision–Language Model (VLM) \texttt{GPT-6 Astra}~\citep{gpt6astra2026}, and VLM-based planners with oracle access to object states. Our empirical results show that pre-trained VLAs leverage their prior knowledge to outperform from-scratch models in low-data settings (50 demonstrations), yet still exhibit substantial difficulty with tasks requiring complex material interactions and assembly-level reasoning. Without any fine-tuning, the VLM-based policy reaches roughly three times the success rate of the best VLA, but its success still drops by about a fifth on mutated tasks, and some tasks remain nearly unsolved by all policies. Even with oracle state access, modular planners driven by VLMs achieve reasonable performance on seed tasks but fail to generalize effectively to mutated task variants. Overall, \papername offers a rigorous framework for quantifying the gap between low-level manipulation proficiency and high-level cognitive adaptation.

In summary, our contributions are as follows:
\begin{itemize}
    \setlength{\itemsep}{-1pt}
    \item We introduce \papername, a new robotic benchmark that evaluates cognitive reasoning, creative tool use, and robustness to unexpected challenges in bi-manual manipulation.
    \item We propose an automated, multi-agent task generation pipeline that enables scalable construction of diverse, reasoning-centric manipulation tasks.
    \item We benchmark existing robotic policy methods, revealing achievements and limitations in reasoning and adaptation beyond low-level skill execution.
\end{itemize}

\section{Related Works}
\label{sec:related}

\begin{table}[t]
\centering
\caption{\textbf{Benchmark Comparison.} We compare \papername{} against related works. \papername{} features an automatic generation pipeline and provides systematic support for tasks involving diverse materials (fluids, soft bodies), creative tool use, and strategy adaptation.}
\label{tab:benchmark_comparison}
\resizebox{\textwidth}{!}{%
\begin{tabular}{llcccc ccc}
\toprule
\multirow{3}{*}{\textbf{Benchmark}} & \multirow{3}{*}{\textbf{Task Type}} & \multirow{3}{*}{\textbf{Curation}} & \multirow{3}{*}{\textbf{Evaluation}} & \multirow{3}{*}{\textbf{\# Tasks}} & \multirow{3}{*}{\textbf{Horizon}} & \multicolumn{3}{c}{\textbf{Task Features}} \\
\cmidrule(lr){7-9}
& & & & & & \textbf{Diverse} & \textbf{Tool} & \textbf{Strategy} \\
& & & & & & \textbf{Materials} & \textbf{Use} & \textbf{Adaptation} \\
\midrule
RLBench \cite{james2020rlbench} & Unimanual & Human & Binary Success & 100 & Short--Long & \xmark & \cmark & \xmark \\
LIBERO \cite{liu2023libero} & Unimanual & Human & Binary Success & 130 & Short--Long & \xmark & \xmark & \xmark \\
RoboTwin 2.0 \cite{chen2025robotwin} & Bi-manual & Human & Binary Success & 50 & Short--Med & \xmark & \xmark & \xmark \\
RoboEval \cite{wang2025roboeval} & Bi-manual & Human & Succ. + Score & 10 & Short--Med & \xmark & \xmark & \xmark \\
VLABench \cite{zhang2025vlabench} & Unimanual & Human & Binary Success & 100 & Short--Long & \xmark & \cmark & \xmark \\
MacGyver \cite{tian2024macgyver} & Text Planning & Generated & Human Judge & 1.6k & - & \cmark & \cmark & \cmark \\
\midrule
\textbf{RoboWits (Ours)} & \textbf{Bi-manual} & \textbf{Generated} & \textbf{Succ. + Score} & \textbf{200} & \textbf{Short--Long} & \textbf{\cmark} & \textbf{\cmark} & \textbf{\cmark} \\
\bottomrule
\end{tabular}%
}
\end{table}

\subsection{Robotic Manipulation and Benchmarks}
\label{subsec:robotic_benchmark}

Robot policy model has seen rapid progress recently due to increased data available and better architectures~\cite{brohan2022rt,zitkovich2023rt, liu2024rdt, kim2024openvla, li2024cogact, team2024octo, ye2024latent, wang2024rise, chi2025diffusion, fu2024mobile, wen2025tinyvla, lee2025molmoact, ke20243d, qiu2025lucibot}. With modern physics simulators such as Genesis \cite{authors2024genesis}, SAPIEN~\cite{xiang2020sapien}, MuJoCo~\cite{todorov2012mujoco}, IsaacGym~\cite{makoviychuk2021isaac}, and other specialized simulators~\cite{wang2023softzoo,lin2024ubsoft,song2025oceansim,cao2026dlo}, many robotic benchmarks have been developed and have played an important role in evaluating and improving robot policy models~\cite{gu2023maniskill2, pumacay2024colosseum, li2024evaluating, chernyadev2024bigym, mees2022calvin, chen2025benchmarking}. 
RLBench~\cite{james2020rlbench} includes 100 diverse human-designed manipulation tasks, LIBERO~\cite{liu2023libero} features task suites across four dimensions, RoboTwin 2.0~\cite{chen2025robotwin} provides 50 bi-manual manipulation tasks, RoboEval~\citep{wang2025roboeval} incorporates structured evaluation, and VLABench~\citep{zhang2025vlabench} provides manipulation tasks requiring language-level reasoning. Other benchmarks such as HomeRobot~\cite{yenamandra2023homerobot}, RoboCasa~\cite{nasiriany2024robocasa}, and RoboSuite~\cite{zhu2020robosuite} further extend toward mobile manipulation. 
Despite their success, most existing benchmarks mainly emphasize skill execution under well-specified settings.  Our benchmark and automated pipeline take a step further by systematically generating and evaluating bi-manual tasks that target explicit reasoning, diverse materials, creative tool use, and robust adaptation under unexpected constraints.

\subsection{Creative Problem Solving}

Creative problem solving has been widely studied in domains such as NLP and cognitive science, where benchmarks evaluate multi-step planning, strategy revision, and compositional generalization~\cite{guilford1967creativity,kaufman2009beyond,tian2024macgyver,collins2022structured,wang2023newton}. However, these benchmarks are often limited to specific settings like text-only and require extensive human evaluation. 
In robotics, physics simulators provide a natural testbed to study creative problem solving through grounded interaction and automatic success metrics. Some robotic benchmarks begin to touch on related aspects, often through multi-task instruction following or language-conditioned manipulation~\cite{yu2020meta,mees2022calvin,geng2025roboverse,zhang2025vlabench}, but they do not systematically isolate and evaluate creative tool use and adaptation under unexpected constraints. 
In contrast, \papername targets these capabilities with reasoning-centric bi-manual tasks and metric-based evaluation at scale.

\subsection{Automated Task Generations}

Automatically generating tasks is essential for scaling benchmarks while reducing the cost of manual design and human labor. 
Prior work explores \textit{task diversification} through domain randomization~\cite{james2020rlbench,chen2025robotwin,zhouharp}, which mainly varies superficial factors such as instructions, textures, and object placements.
Several lines of research explore \textit{scene generation} via procedural generation with hand-crafted rules~\cite{deitke2022,liu2023libero}, image-driven pipelines that leverage 2D priors to construct 3D layouts~\cite{hao2025mesatask,wang2024architect}, and, more recently, LLM-based methods that either generate symbolic constraints for external solvers or directly predict object placements through prompting~\cite{wang2023robogen,yang2024holodeck}.
Other works focus on \textit{metric generation}, where task success or even rewards are defined through programmatic checks over simulator states and physical conditions~\cite{huang2023diffvl, ma2024dreureka, lin2025robotsmith}. 
Building on these directions, we propose a unified multi-agent pipeline that jointly performs task creation, diversification, verification, scene instantiation, and metric synthesis. This enables scalable generation of reasoning-centric bi-manual tasks with diverse solution spaces, varying cognitive difficulty, and ground-truth evaluation.
\section{\papername Benchmark}
\label{sec:method}

We first introduce our task settings in Sec.~\ref{sec:task_settings}, then our automatic agentic task generation pipeline in Sec.~\ref {sec:taskgen}, and finally present the benchmark in Sec.~\ref {sec:benchmark}.

\subsection{Task Settings}
\label{sec:task_settings}

To evaluate robotic creative problem-solving and strategy adaptation capabilities under unexpected conditions, \papername tasks are designed to tightly couple cognitive reasoning with physical interactions. All tasks are set in a tabletop environment with a dual-arm robot equipped with parallel grippers. The workspace contains everyday objects with diverse shapes, sizes, and material properties. Each task $T$ is defined by a natural language instruction $\mathcal{I}$, an object list $\mathcal{O}$, an initial 3D scene configuration $\mathcal{C}$, an evaluation metric $\mathcal{M}$, and a difficulty score $s \in \{1,2,3,4,5\}$.
$$T=(\mathcal{I, O, C, M}, s)$$
The instruction $\mathcal{I}$ specifies only the goal and intentionally avoids hinting at the solution. For example, an instruction may state \textit{``pinch the bank card"} rather than \textit{``push the bank card to the table edge and pick it up"}. The reasoning part to discover an appropriate solution is left entirely to the robot. Each object in the object list $\mathcal{O}$ contains a natural language description of the desired \textit{appearance} and \textit{functional} attributes, while $\mathcal{C}$ instantiates all the objects with a specified 3D model, parameterized physical material, and group relationships. 
The evaluation metric $\mathcal{M}$ includes both a binary success indicator and a continuous progress score function. The difficulty score $s$ reflects both physical execution difficulty and reasoning complexity.

\subsection{Task Generation}
\label{sec:taskgen}

\begin{figure}
    \centering
    \includegraphics[width=\linewidth]{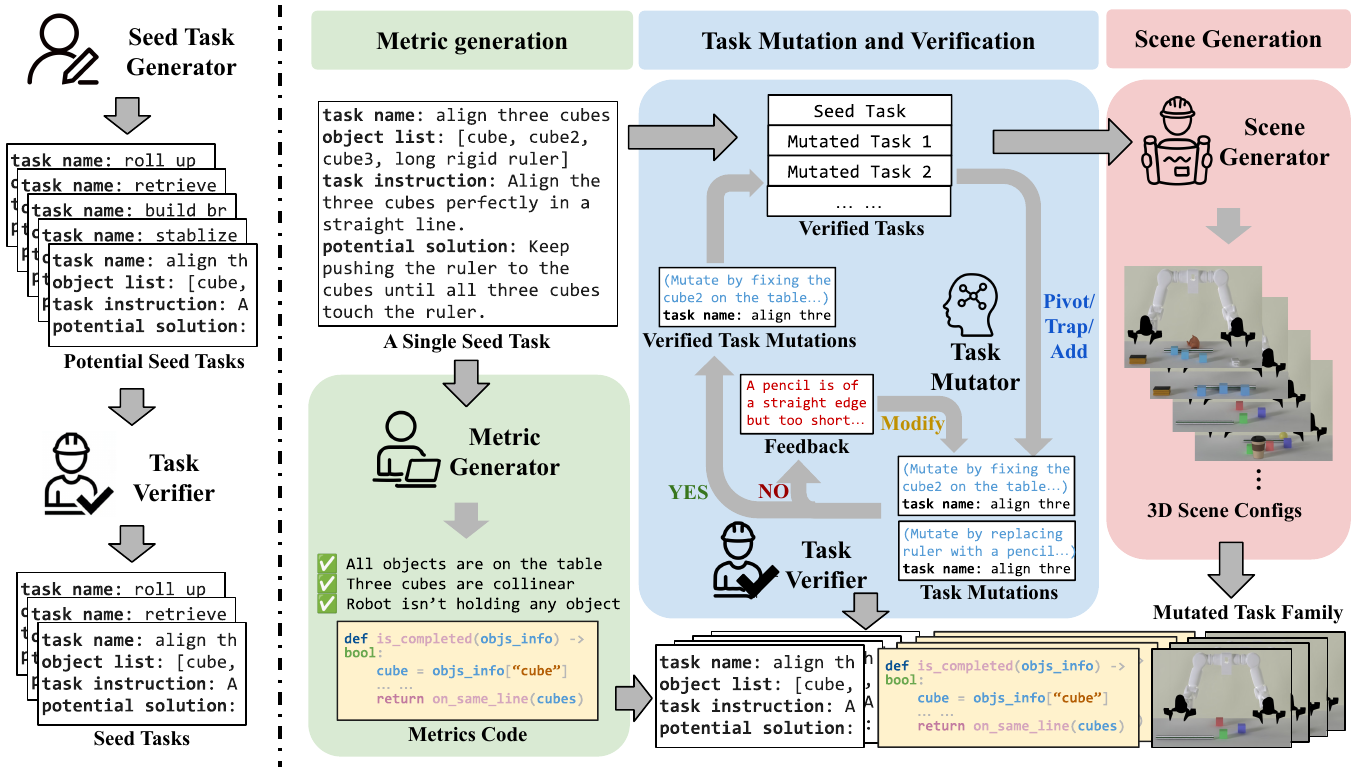}
    \caption{Overview of our automated task generation pipeline, where foundation model-powered agents (depicted by role-based avatars) collaborate to design, validate, and instantiate tasks. Each component with a human-like icon is an agent powered by a foundation model. \textbf{Left:} seed task generation and verification at the specification (verbal) level. \textbf{Right:} a verified seed task is expanded into diverse creative problem-solving task variants through iterative mutation and verification, and is then instantiated with 3D scene configurations and executable evaluation metrics in the simulator to produce the final benchmark tasks.}
    \label{fig:task_gen_pipeline}
\end{figure}

To generate diverse and cognitively challenging tasks at scale, we propose a multi-agent task generation pipeline as illustrated in Figure~\ref{fig:task_gen_pipeline}. The system consists of 5 distinct components: a Seed Task Generator, a Task Verifier, a Metric Generator, a Task Mutator, and a Scene Generator. Each component acts as an independent agent powered by a foundation model, collaborating to complete the task generation process. We describe these components in detail below and provide an illustrative example in Appendix~\ref{sec:case}.

\subsubsection{Seed Task Generation}
\label{subsec:seed_task}

We define three task taxonomies: geometry-based reasoning, material-based reasoning, and assembly reasoning. These categories require robots to reason about how geometric or material properties relate to object functionality, as well as how multiple objects can be composed into a functional tool or system to accomplish a goal.
To balance structure and diversity, we represent all three categories in a unified task schema, which we provided an example of in Appendix~\ref{sec:task_schema}. The schema standardizes task specification and evaluation-relevant fields, while still allowing substantial variation in task content, scene composition, and difficulty.
Each seed task consists of: (i) a natural-language instruction, (ii) a structured object list, (iii) a potential solution (used only for verification and not exposed to the robot), and (iv) a short description explaining why the task requires reasoning. 

\begin{wrapfigure}{r}{0.5\textwidth}
\begin{minicodebox}{json}{}
{
  "object_name": "container",
  "appearance_attribute": ["tall", "thin", "transparent", "opening too narrow for gripper"],
  "functional_attribute": ["rigid", "movable", "stable"],
  "potential_instances": ["a high thin cup", "a bottle with narrow mouth", "a tall glass vase"]
}
\end{minicodebox}
\caption{An example object in the task schema.}
\label{fig:object}
  
\end{wrapfigure}

As shown in Fig~\ref{fig:object}, each item in the object list is defined by its appearance attributes, functional attributes, and a set of candidate instances. This abstraction encourages the Seed Task Generator to focus on an object’s underlying role in the task, rather than overfitting to a specific instance. For example, an object may be designated by the abstract name \textit{container}, with appearance attributes indicating a narrow opening, and functional attributes specifying a rigid material and stable placement on the table. The candidate instances then help concretize the abstraction by mapping it to everyday objects. This representation supports generating diverse task instances that share the same reasoning pattern but differ in concrete object realizations.

\subsubsection{Task Verification}
\label{subsec:verification}

The Task Verifier is a tool-augmented and simulator-aware agent. It determines task validity along three dimensions: simulatability, solution feasibility, and solution efficiency.

\textbf{Simulatability}. The verifier first checks whether the task can be instantiated and executed within the simulator. Tasks that require physical properties or interactions beyond the simulator’s capabilities are immediately rejected. To ensure that all referenced objects can be created in the simulator, the verifier can invoke an asset retrieval tool to search for suitable 3D assets from both a local asset library and online resources\footnote{https://www.blenderkit.com/}, and a 3D asset generation tool to try to create the assets. If an object cannot be matched to an appropriate 3D asset and fails to be generated, the task cannot be instantiated and is therefore rejected.

\textbf{Solution feasibility and efficiency}. Given the potential solution, the verifier examines whether the required operations are realistically executable, rejecting solutions that rely on extreme precision control, highly dynamic motions, or operations beyond the capability of a bi-arm robot with parallel grippers. In addition, the verifier checks whether the task admits significantly easier alternative solutions. Tasks with such shortcut solutions are rejected to ensure that success requires non-trivial reasoning. A human validation study (Appendix~\ref{sec:app_human_study}) confirms that sampled tasks are solvable by teleoperation and largely free of shortcuts.

\textbf{Difficulty score}. For tasks that pass the above checks, the verifier assigns an operational difficulty score $s_o$ based on execution-related factors, including the number of required steps, the complexity of the manipulation primitives (e.g., simple pick-and-place versus precise control), and whether bi-manual coordination is required. We validate the difficulty scores against measured policy performance in Appendix~\ref{sec:app_difficulty}.

\begin{figure}
    \centering
    \includegraphics[width=\linewidth]{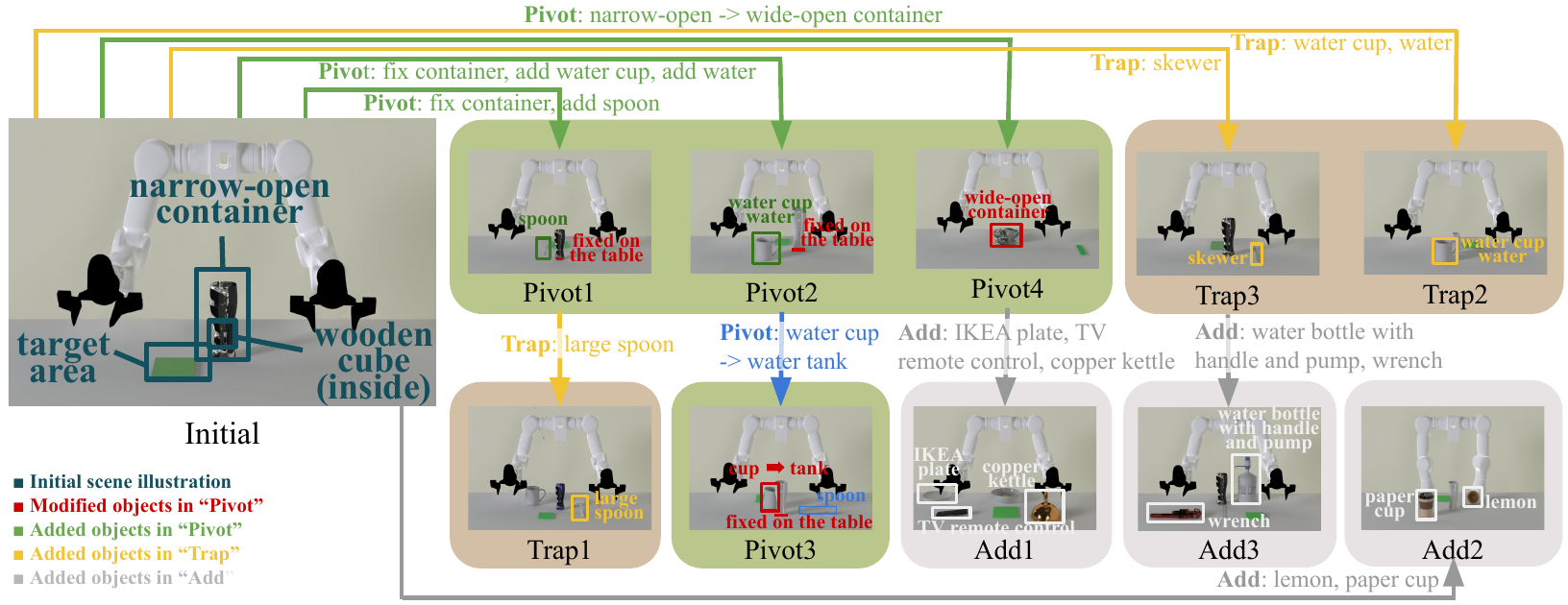}
    \caption{Illustration of mutation on the \textit{retrieve cube} seed task. Starting from the initial seed task, the mutator generates diverse task variations via \textit{pivot}, \textit{trap}, and \textit{add} mutations, resulting in a family of tasks with distinct challenges and diverse difficulty.}
    \label{fig:mutation_retrive_cube}
\end{figure}

\subsubsection{Task Mutation}
\label{subsec:mutation}

Given a seed task, the \textbf{Task Mutator} agent collaborates with the \textbf{Task Verifier} agent to generate a set of diverse and valid task mutations that require a novel solution by adding minimal constraints. The mutator proposes candidate mutations by modifying the task specification (primarily the object list and object attributes), while preserving the original task goal and evaluation metric. The verifier then checks their validity and provides feedback when revisions are needed. Through this iterative process, we obtain a family of task instances with diverse challenges and graded difficulty. To systematically increase task diversity and difficulty, we design three mutation strategies. 

\textbf{Pivot} strategy blocks the current solution by removing or modifying key objects in the object list, and introduces new objects that enable alternative valid solutions. This mutation promotes diversity in feasible solutions, rather than generating superficial variations.

\textbf{Trap} strategy increases both reasoning and execution difficulty by injecting "trap objects" into the scene. Trap objects are designed to appear useful but are functionally ineffective due to subtle constraints such as mismatched resizing, being fixed to the table, being placed out of reach, or having unexpected material properties. For example, a board that visually resembles a rigid lever can be assigned a soft material, making it unsuitable for applying torque. These traps substantially increase reasoning difficulty by introducing deceptive failure modes and test whether a policy can recognize failure and recover or adapt its strategy accordingly.

\textbf{Add} strategy increases scene complexity by adding useless objects as distractors. Although these objects don't affect task solutions, they introduce clutter and perceptual noise, making tool selection and reasoning more challenging.

\textbf{Mutation-Verification Workflow} The overall mutation-verification workflow is summarized as Algorithm~\ref{alg:mutate_and_verify} in the Appendix~\ref{sec:mut_and_ver}. We maintain a \textit{valid task pool}, initialized with the seed task. At each mutation step, we randomly sample a task from the pool and select a mutation strategy from \{\emph{pivot}, \emph{trap}, \emph{add}\}. The mutator then generates a candidate task variant for the verifier to evaluate. If the candidate is deemed invalid, the verifier provides structured feedback, and the mutator revises the task accordingly. This back-and-forth refinement is allowed for up to 3 rounds. Verified task variants are added to the pool, and others are discarded. To avoid repetitive variants, we prevent applying the same mutation strategy to a task more than twice.

In practice, we prioritize the \textit{pivot} strategy in the early stages to maximally explore the solution space while keeping tasks minimal (i.e., avoiding redundant or distracting objects). This step is critical for discovering fundamentally different solution strategies rather than superficial variations.  
After establishing diverse solution families, the \textit{trap} strategy introduces deceptive but inefficient alternatives that increase cognitive difficulty. The \textit{add} strategy is applied to increase scene clutter, primarily raising execution and control difficulty without changing the core reasoning structure. 
We show example mutation-verification trees in Figure~\ref{fig:task_mutation_trees} in the Appendix.

\definecolor{success_node}{RGB}{76,175,80}
\definecolor{failed_node}{RGB}{239,83,80}

\subsubsection{Task Scene Generation}
\label{subsec:scene_gen}

Given a task specification, the Scene Generator agent is invoked to produce the initial 3D scene configuration, following prior work~\citep{hu2024scenecraft, zhang2026gs}. This includes automatically retrieving new 3D asset models if necessary, assigning physical properties (e.g., material types, density, and friction coefficients), and determining the initial positions and orientations of all objects. The agent is equipped with several tools to interact with the physics simulator, iteratively place objects, assign physical parameters, and render four views of the current scene and the scene after some physics steps to verify that the generated scene code is plausible and that the scene is stable for robots to execute tasks. The Scene Generator agent enables scalable, diverse scene instantiation while maintaining both physical and visual realism.

\begin{figure}[t]
    \centering
    \includegraphics[width=\linewidth]{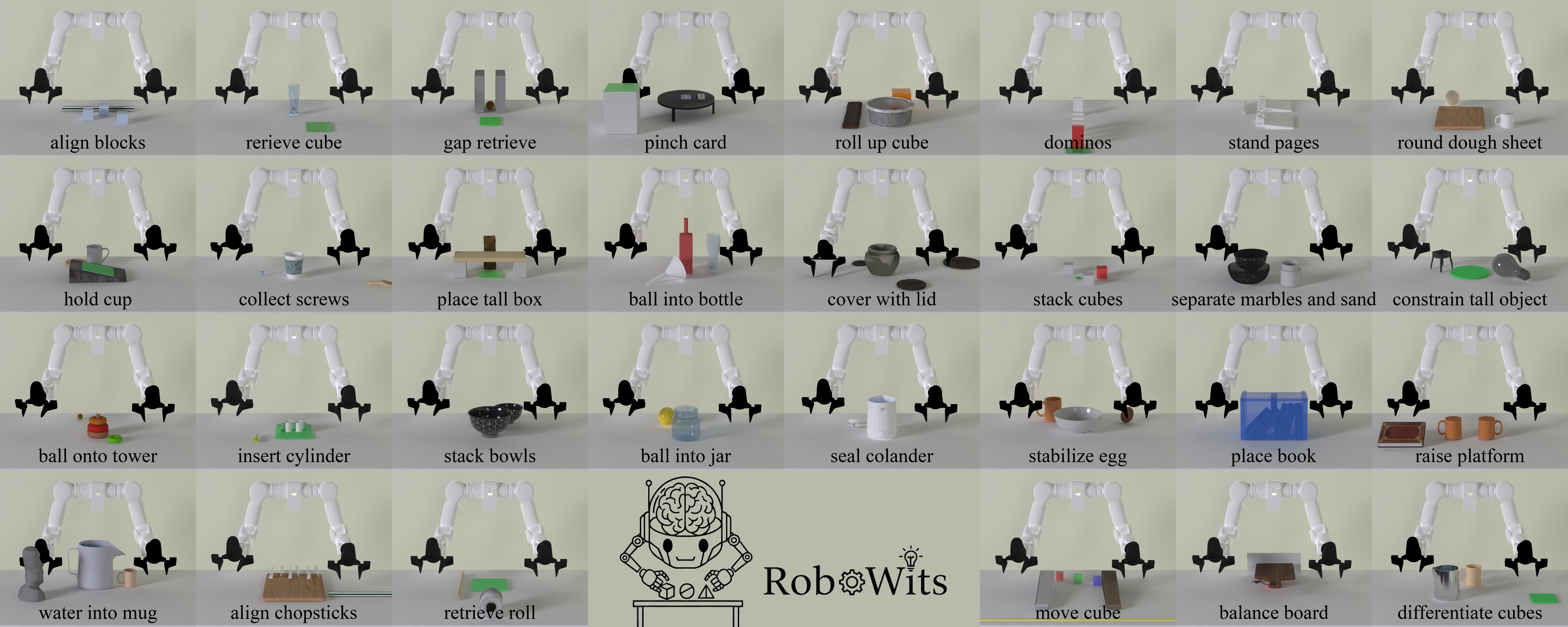}
    \caption{Gallery of the 30 bi-manual creative problem-solving manipulation tasks in \papername.}
    \label{fig:task_gallery}
\end{figure}

\subsubsection{Task Metric Generation}
\label{subsec:metric_gen}

The final stage of the pipeline generates task evaluation metrics, including a binary success indicator and a detailed progress score function. Following prior work~\citep{ma2024dreureka, lin2025robotsmith}, we predefine a set of simulator APIs for querying object states and physical conditions, which the Metric Generator agent uses to produce executable evaluation code. Crucially, all mutated task variations share the same goal and thus the same evaluation metric of their seed task. This means metric generation occurs only once per seed task. While manual verification remains a standard necessity across benchmarks to mitigate edge cases, our design drastically reduces this scaling bottleneck. By amortizing evaluation logic across mutation task families, humans verify just 30 unique scripts to guarantee reliable high-fidelity evaluation for the entire 200-task benchmark with minimal labor.

\subsection{Benchmark Overview}
\label{sec:benchmark}

\begin{figure}[t]
    \centering
    
    \begin{minipage}[b]{0.65\textwidth}
        \centering
        \includegraphics[width=\linewidth]{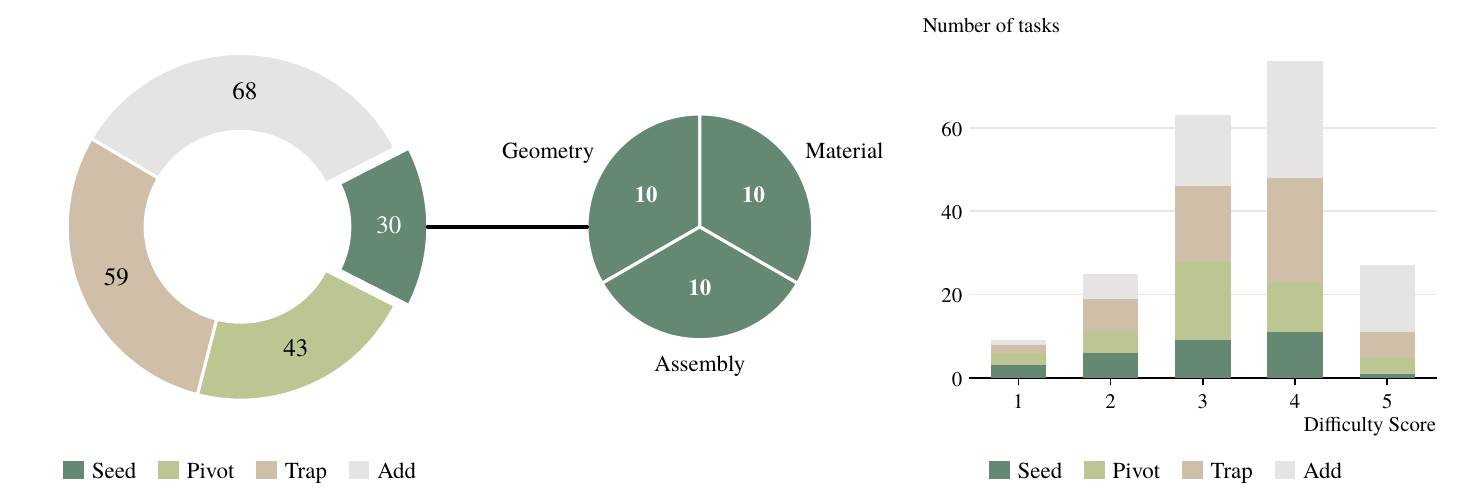}
        \caption{(Left) Distribution of task types and the three task categories in the 30 seed tasks. (Right) Difficulty distribution of the 200 tasks.}
        \label{fig:task_statistics}
    \end{minipage}
    \hfill
    \begin{minipage}[b]{0.34\textwidth}
        \centering
        \begin{tabular}{lccc}
        \toprule
         & Pivot & Trap & Add \\
        \midrule
        count & 43 & 59 & 68 \\
        mean & 0.14 & 0.42 & 0.82 \\
        max & 3 & 2 & 2 \\
        min & -2 & 0 & -1 \\
        std & 0.91 & 0.65 & 0.71 \\
        \bottomrule
        \end{tabular}
        \captionof{table}{Changes in difficulty scores ($\Delta$) across different task mutation strategies.}
        \label{tab:delta_summary}
    \end{minipage}
\end{figure}

\papername contains $200$ diverse bi-manual manipulation tasks generated by our automated agentic pipeline introduced in Sec.~\ref{sec:taskgen}, featuring $30$ seed tasks and $2\sim8$ mutations each requiring novel solutions, spanning three categories: geometry-based reasoning, material-based reasoning, and assembly reasoning, as shown in Fig.~\ref{fig:task_gallery}. Each task is assigned a difficulty score ranging from 1 to 5, reflecting increasing levels of reasoning and execution complexity, as shown in Fig.~\ref{fig:task_statistics}. Table~\ref{tab:delta_summary} further demonstrates that the mutation strategies successfully adjust task difficulty as intended. Each task captures a distinct problem-solving pattern and is paired with a fixed evaluation metric that precisely defines task success. We instantiate all the tasks in Genesis~\cite{authors2024genesis}, a unified physics simulator that supports diverse materials and robot control. Coverage statistics over objects, spatial relations, and manipulation primitives are listed in Appendix~\ref{sec:app_coverage}. 

To support training and fine-tuning, we collect and release human teleoperation demonstrations for \papername. For the 30 seed tasks, we provide demonstrations from simulated teleoperation. We analyze the diversity of these demonstrations, the manipulation skills they cover, and the spread of initial object placements in Appendices~\ref{sec:app_demo_diversity}, \ref{sec:app_skill_coverage}, and~\ref{sec:app_scene_variation}, respectively. 
\section{Experiments}
\label{sec:exp}

We conduct a series of experiments on \papername to answer four core research questions.
\begin{itemize}
    \setlength{\itemsep}{-1pt}
    \item RQ1: Can Vision-Language-Action models control robots to creatively solve manipulation problems?
    \item RQ2: Can Vision-Language-Action models perform robustly in the face of unexpected challenges?
    \item RQ3: Can a policy built on a frontier Vision-Language Model close this gap without fine-tuning?
    \item RQ4: Can Vision Language Model-driven modular planners think out of the box?
\end{itemize}

\subsection{Experimental Setup}

We select 10 representative task families and report results on \textbf{RoboWits-10} in the main paper, and report results on the full benchmark in Appendix~\ref{sec:app_full_suite}. While the capability to creatively solve manipulation problems is arguably learnable by imitation, and our benchmark demands zero-shot evaluation ideally, no existing robotic models except policies built upon advanced foundation models like \texttt{Astra} have demonstrated such generalization. We follow similar principles in robotics benchmarks~\citep{james2020rlbench, liu2023libero} to fine-tune/train robotic models in a multi-task setting with 50 collected human demonstrations on seed tasks, and test them on the seed tasks for 50 trials, and each of the mutation tasks for 20 trials, respectively.\footnote{\texttt{Astra-Policy} is evaluated for 10 trials for each seed task and 5 trials for each mutation task due to cost constraints.} All trials have randomized object positions and rotations.

\textbf{Robot system details.} \papername uses two 7-DoF Marvin arms with a parallel gripper and three cameras (two wrist and one base). While \papername supports a flexible control interface, we conduct experiments with Cartesian space control unless otherwise stated.

\textbf{Baselines.} We evaluate robot policy models, pre-trained VLAs, zero-shot VLM-based policies, and Oracle VLM-based planners.
\begin{itemize}[leftmargin=*, itemsep=-3mm]
    \setlength{\itemsep}{0pt}
    \item \texttt{ACT}~\citep{zhao2023learning}, a transformer-based policy that learns action chunking from demonstrations
    \item \texttt{$\pi_0$}~\citep{black2026pi0visionlanguageactionflowmodel}, a pre-trained 3B VLA that predicts continuous action with flow matching.
    \item \texttt{$\pi_{0.5}$}~\citep{black2025pi_}, a 3B VLA pre-trained with internet-scale data and robotics data that can generalize better.
    \item \texttt{Astra-Policy}, a policy built on the VLM \texttt{GPT-6 Astra}~\citep{gpt6astra2026}, evaluated zero-shot without fine-tuning on our demonstrations. It shares the same interface as the VLAs above: it receives the language instruction and the same camera observations and robot states as input, and outputs actions in the same 16-dimensional action space.
    \item \texttt{VLM Planner$^{**}$} A modular planner with oracle access to object states and perfect action primitives. It employs a VLM (\texttt{gpt-4o}) for high-level planning and assumes error-free execution.
    \item \texttt{VLM Controller$^*$} A modular planner similar to the above, but without perfect action primitives. It uses the same VLM for planning but relies on a human-scripted low-level controller for execution.
\end{itemize}

\subsection{Experimental Results}

\begin{table}[t]
\centering
\caption{\textbf{Results on 10 Seed Tasks and their Mutations on \papername.} Rows are split by Method and Task Type (Seed/Mutation). Columns report Success Rate (SR) and averaged Progress Score (PS) over 50 trials for each seed task and 20 trials for each mutation task. Objects are randomly placed per trial. * Models assume oracle state. ** Models assume perfect action primitives and do not have Progress Score. Best performance per group is \textbf{bolded}.}
\label{tab:main}
\resizebox{\textwidth}{!}{%
\begin{tabular}{llcccccccccccccccccccccc}
\toprule
& & \multicolumn{10}{c}{\textbf{Geometry Tasks}} 
& \multicolumn{4}{c}{\textbf{Assembly Tasks}} 
& \multicolumn{6}{c}{\textbf{Material Tasks}} 
& \multicolumn{2}{c}{\textbf{All}} \\
\cmidrule(lr){3-12} \cmidrule(lr){13-16} \cmidrule(lr){17-22} \cmidrule(lr){23-24}

& & \multicolumn{2}{c}{\textbf{Align Blocks}} & \multicolumn{2}{c}{\textbf{Retr. Cube}} & \multicolumn{2}{c}{\textbf{Dominos}} & \multicolumn{2}{c}{\textbf{Gap Retr.}} & \multicolumn{2}{c}{\textbf{Stand Bulb}} 
& \multicolumn{2}{c}{\textbf{Cover Lid}} & \multicolumn{2}{c}{\textbf{Ball Tower}} 
& \multicolumn{2}{c}{\textbf{Hold Cup}} & \multicolumn{2}{c}{\textbf{Pinch Card}} & \multicolumn{2}{c}{\textbf{Water Mug}} 
& \multicolumn{2}{c}{\textbf{Avg.}} \\

\cmidrule(lr){3-4} \cmidrule(lr){5-6} \cmidrule(lr){7-8} \cmidrule(lr){9-10} \cmidrule(lr){11-12} 
\cmidrule(lr){13-14} \cmidrule(lr){15-16} 
\cmidrule(lr){17-18} \cmidrule(lr){19-20} \cmidrule(lr){21-22} 
\cmidrule(lr){23-24}

\textbf{Method} & \textbf{Type} & \textbf{SR} & \textbf{PS} & \textbf{SR} & \textbf{PS} & \textbf{SR} & \textbf{PS} & \textbf{SR} & \textbf{PS} & \textbf{SR} & \textbf{PS} & \textbf{SR} & \textbf{PS} & \textbf{SR} & \textbf{PS} & \textbf{SR} & \textbf{PS} & \textbf{SR} & \textbf{PS} & \textbf{SR} & \textbf{PS} & \textbf{SR} & \textbf{PS} \\
\midrule

\multicolumn{24}{l}{\textbf{Vision-Language-Action Models}} \\
\midrule

\multirow{2}{*}{\texttt{ACT}} 
& Seed & 14.0\% & 0.61 & 2.0\% & 0.29 & 52.0\% & 0.87 & 2.0\% & 0.51 & 0.0\% & 0.54 & 0.0\% & 0.48 & 0.0\% & 0.55 & 0.0\% & 0.53 & 0.0\% & 0.41 & 2.0\% & \textbf{0.36} & 7.2\% & 0.52 \\
& Mut. & 10.0\% & 0.58 & 0.0\% & \textbf{0.32} & 38.3\% & 0.86 & 0.0\% & 0.43 & 0.0\% & 0.41 & 0.0\% & 0.37 & 0.0\% & 0.48 & 0.0\% & 0.39 & 0.0\% & 0.53 & \textbf{0.7\%} & 0.43 & 4.9\% & 0.48 \\
\midrule

\multirow{2}{*}{\texttt{$\pi_0$}}
& Seed & 14.0\% & \textbf{0.63} & \textbf{6.0\%} & \textbf{0.30} & \textbf{86.0\%} & \textbf{0.97} & 18.0\% & 0.65 & 0.0\% & \textbf{0.55} & 0.0\% & 0.53 & 0.0\% & 0.62 & 0.0\% & \textbf{0.66} & \textbf{10.0\%} & \textbf{0.63} & 0.0\% & \textbf{0.36} & \textbf{13.4\%} & \textbf{0.59} \\
& Mut. & \textbf{20.8\%} & 0.62 & 0.8\% & 0.31 & \textbf{57.5\%} & \textbf{0.87} & 5.0\% & 0.46 & 0.0\% & \textbf{0.42} & 0.0\% & 0.43 & 0.0\% & 0.54 & 0.0\% & 0.45 & \textbf{9.0\%} & \textbf{0.62} & 0.0\% & 0.45 & \textbf{9.3\%} & 0.52 \\
\midrule

\multirow{2}{*}{\texttt{$\pi_{0.5}$}} 
& Seed & \textbf{20.0\%} & \textbf{0.63} & 0.0\% & 0.28 & 74.0\% & 0.95 & \textbf{20.0\%} & \textbf{0.66} & 0.0\% & \textbf{0.55} & 0.0\% & \textbf{0.58} & 0.0\% & \textbf{0.67} & \textbf{2.0\%} & 0.64 & 2.0\% & 0.54 & \textbf{4.0\%} & 0.35 & 12.2\% & \textbf{0.59} \\
& Mut. & 15.0\% & \textbf{0.63} & \textbf{1.7\%} & \textbf{0.32} & 52.5\% & 0.84 & \textbf{6.0\%} & \textbf{0.49} & 0.0\% & \textbf{0.42} & 0.0\% & \textbf{0.48} & 0.0\% & \textbf{0.58} & 0.0\% & \textbf{0.46} & 7.0\% & 0.59 & 0.0\% & \textbf{0.47} & 8.2\% & \textbf{0.53} \\
\midrule

\multicolumn{24}{l}{\textbf{Vision-Language Model Policy}} \\
\midrule

\multirow{2}{*}{\texttt{Astra-Policy}}
& Seed & 90.0\% & 0.96 & 10.0\% & 0.51 & 70.0\% & 0.92 & 60.0\% & 0.90 & 10.0\% & 0.62 & 60.0\% & 0.90 & 0.0\% & 0.65 & 10.0\% & 0.90 & 30.0\% & 0.91 & 10.0\% & 0.30 & 35.0\% & 0.76 \\
& Mut. & 86.7\% & 0.90 & 24.0\% & 0.43 & 76.7\% & 0.95 & 20.0\% & 0.59 & 0.0\% & 0.50 & 20.0\% & 0.73 & 0.0\% & 0.55 & 16.0\% & 0.68 & 28.0\% & 0.87 & 5.7\% & 0.48 & 27.7\% & 0.67 \\
\midrule

\multicolumn{24}{l}{\textbf{Modular Planners}} \\
\midrule

\multirow{2}{*}{\texttt{VLM Planner$^{**}$}} 
& Seed & \textbf{52.0\%} & - & \textbf{94.0\%} & - & 0.0\% & - & \textbf{80.0\%} & - & \textbf{78.0\%} & - & \textbf{80.0\%} & - & \textbf{100.0\%} & - & \textbf{100.0\%} & - & \textbf{24.0\%} & - & 0.0\% & - & \textbf{60.8\%} & - \\
& Mut. & \textbf{33.3\%} & - & \textbf{20.0\%} & - & 0.0\% & - & \textbf{80.0\%} & - & \textbf{42.9\%} & - & \textbf{80.0\%} & - & \textbf{87.5\%} & - & \textbf{100.0\%} & - & 0.0\% & - & 0.0\% & - & \textbf{44.4\%} & - \\
\midrule

\multirow{2}{*}{\texttt{VLM Controller$^*$}} 
& Seed & 0.0\% & 0.00 & 20.0\% & 0.71 & 0.0\% & 0.61 & 0.0\% & 0.64 & 30.0\% & 0.56 & 0.0\% & 0.37 & 10.0\% & 0.35 & 10.0\% & 0.42 & 0.0\% & 0.55 & 0.0\% & 0.29 & 7.0\% & 0.45 \\
& Mut. & \textbf{33.3\%} & 0.50 & 0.0\% & 0.38 & 0.0\% & 0.32 & 0.0\% & 0.76 & 14.3\% & 0.12 & 0.0\% & 0.41 & 0.0\% & 0.33 & 0.0\% & 0.49 & 0.0\% & 0.00 & 0.0\% & 0.26 & 4.8\% & 0.36 \\

\bottomrule
\end{tabular}%
}
\end{table}

\textbf{Current VLAs struggle on \papername even with single-task fine-tuning.} As detailed in Table~\ref{tab:main}, even after fine-tuning on 50 human demonstrations, pre-trained models like \texttt{$\pi_0$} exhibit only preliminary success on seed tasks. Full results on all 30 seed tasks and their mutations are reported in Appendix~\ref{sec:app_full_suite}, and repeated evaluations with standard errors are reported in Appendix~\ref{sec:app_repeated_eval}. In contrast, imitation learning baselines trained from scratch, such as \texttt{ACT}, fail significantly; this highlights the difficulty of our tasks and the insufficiency of limited demonstrations for models lacking strong vision-language pre-training.

\textbf{VLAs lack robustness under unexpected challenges.} A comparison between seed tasks and their corresponding mutations reveals a consistent performance drop across all VLAs, with average success rates falling by roughly one third (e.g., from $13.4\%$ to $9.3\%$ for \texttt{$\pi_0$}, and from $12.2\%$ to $8.2\%$ for \texttt{$\pi_{0.5}$}). For instance, in the \textit{Dominos} task, \texttt{$\pi_0$} achieves an $86.0\%$ success rate on the seed version but only $57.5\%$ on its mutations. This gap persists with more demonstrations (Appendices~\ref{sec:app_demo_scaling} and~\ref{sec:app_demo_scaling_ext}) and on a real robot (Appendix~\ref{sec:app_real_world}). This suggests that current VLAs rely on superficial correlations rather than a generalized understanding of task physics. Breakdowns of failures by mutation type and by observed behavior are given in Appendices~\ref{sec:app_mutation_type} and~\ref{sec:app_failure_behavior}.

\textbf{Astra-Policy performs the best but still has large room for improvement.} Sharing the same interaction interface and information, zero-shot \texttt{Astra-Policy} outperforms all fine-tuned VLAs by a wide margin, reaching $35.0\%$ success on seed tasks and $27.7\%$ on mutations, roughly three times that of \texttt{$\pi_0$} ($13.4\%$ and $9.3\%$). It achieves a much higher success rate on easy-to-manipulate tasks like \textit{Align Blocks} ($90.0\%$ on seeds and $86.7\%$ on mutations) and \textit{Cover Lid} ($60.0\%$ on seeds, versus $0.0\%$ for all VLAs), and makes meaningful progress on tasks requiring non-trivial reasoning like \textit{Hold Cup}. Yet tasks like \textit{Ball Tower} and \textit{Stand Bulb} are still nearly unsolved ($10.0\%$ on seeds and $0.0\%$ on mutations), where \texttt{Astra-Policy}  frequently ignores tools such as supporting rings that can make the tasks easier to solve and manipulate the target object, literally leading to catastrophic task failures.

\begin{wrapfigure}{r}{0.4\textwidth}
\centering
        \includegraphics[width=\linewidth]{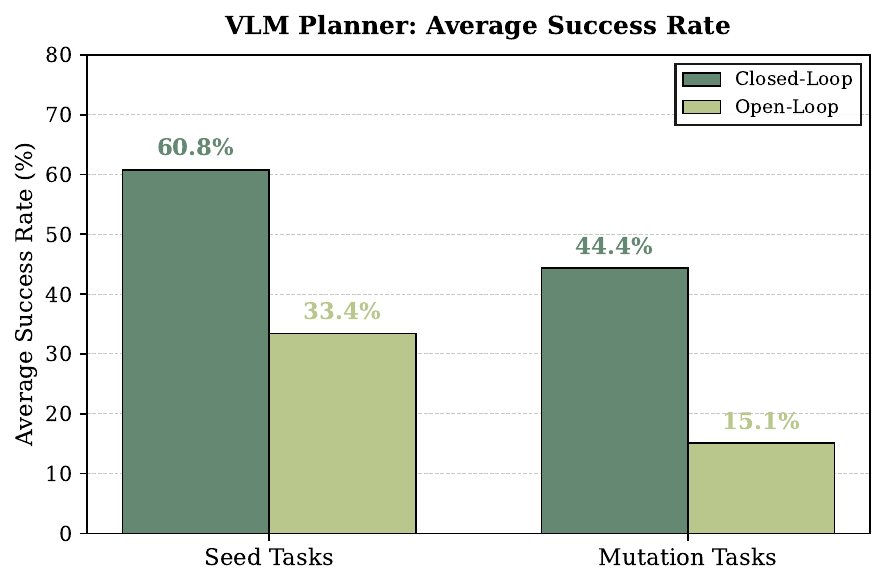}
        \caption{Comparison of \texttt{VLM planner$^{**}$} SRs under closed-loop and open-loop setting.}
        \label{fig:ablate}
\end{wrapfigure}

\textbf{Vision-language planners show promise through closed-loop reasoning.} Given oracle object states and action feedback, VLM-based planners achieve reasonable performance on seed tasks and demonstrate superior robustness to mutations. Our ablation study comparing open-loop and closed-loop variants underscores the necessity of reactivity: closed-loop planning (re-planning after each action) outperforms the open-loop baseline by 27.4 percentage points on seed tasks and 29.3 points on mutations. This gap confirms that real-time feedback is critical for adapting strategies to the unexpected challenges inherent in our benchmark. Results with Gemini 3 Pro as the planner's VLM are reported in Appendix~\ref{sec:app_gemini}.

\textbf{Further findings, along with comprehensive supporting results, are detailed in Appendix~\ref{sec:additional_results}.}
\section{Conclusion}
\label{sec:conclusion}

We present \papername, a new bi-manual manipulation benchmark designed to evaluate robotic creative problem solving, tool use, and robustness to unexpected challenges. To scale task construction beyond manual design, we propose an automated multi-agent pipeline that jointly performs seed task creation, task diversification, verification, scene instantiation, and metric synthesis, enabling diverse reasoning-centric tasks with ground-truth evaluation. Using this pipeline, we build 30 seed tasks and expand them into 200 tasks with graded difficulty across geometry, material, and assembly reasoning. Comprehensive experiments on popular robot policy models in both zero-shot and fine-tuned settings show that while fine-tuning improves basic manipulation skills, substantial gaps remain in reasoning, strategy adaptation, and robustness under deceptive or constrained conditions. We hope \papername will serve as a scalable testbed for developing and diagnosing next-generation reasoning-driven VLA models and robust robotic problem-solving systems.

\section*{Limitations}
\label{sec:limitations}

As a simulation-based benchmark, this work is limited by the sim-to-real gap. Although we use high-fidelity rendering and advanced physics engines to narrow this divide, perfectly replicating complex real-world contact dynamics remains an open challenge. Additionally, the reliance on highly realistic physical interactions yields a computational trade-off: the non-linear dynamics of soft-body simulations are computationally expensive, which limits the rapid execution of large-scale evaluations compared to rigid-body counterparts.

In addition, all models are trained only on successful seed-task demonstrations and never observe a mutated task or a mutation-specific solution. A mutation may require an action that never appears in the seed demonstrations (e.g., \textit{scoop}, Appendix~\ref{sec:app_skill_coverage}), and we do not guarantee that the seed demonstrations cover every skill the mutations require. Low mutation performance may therefore reflect missing action-level supervision as well as limited reasoning. The behavioral failure analysis in Appendix~\ref{sec:app_failure_behavior} cannot fully separate these two causes.

\begin{ack}
We thank Tsun-Hsuan Wang, Yi-Ling Qiao, Pengsheng Guo, and Xiaowen Qiu from Genesis AI for their helpful feedback and insightful discussion. This work was supported by NSF IIS-2441250, NSF IIS-2404386, and MURI.
\end{ack}

\bibliographystyle{abbrv}
\bibliography{main}


\appendix
\newpage
\appendix
\onecolumn

\begin{center}
    {\Large\bfseries Appendix Contents}
\end{center}
\startcontents[appendix]
\begingroup
\small
\printcontents[appendix]{}{1}{\setcounter{tocdepth}{2}}
\endgroup
\clearpage

\section{Task Details}

\begin{figure*}[t]
    \centering
    \includegraphics[width=\linewidth]{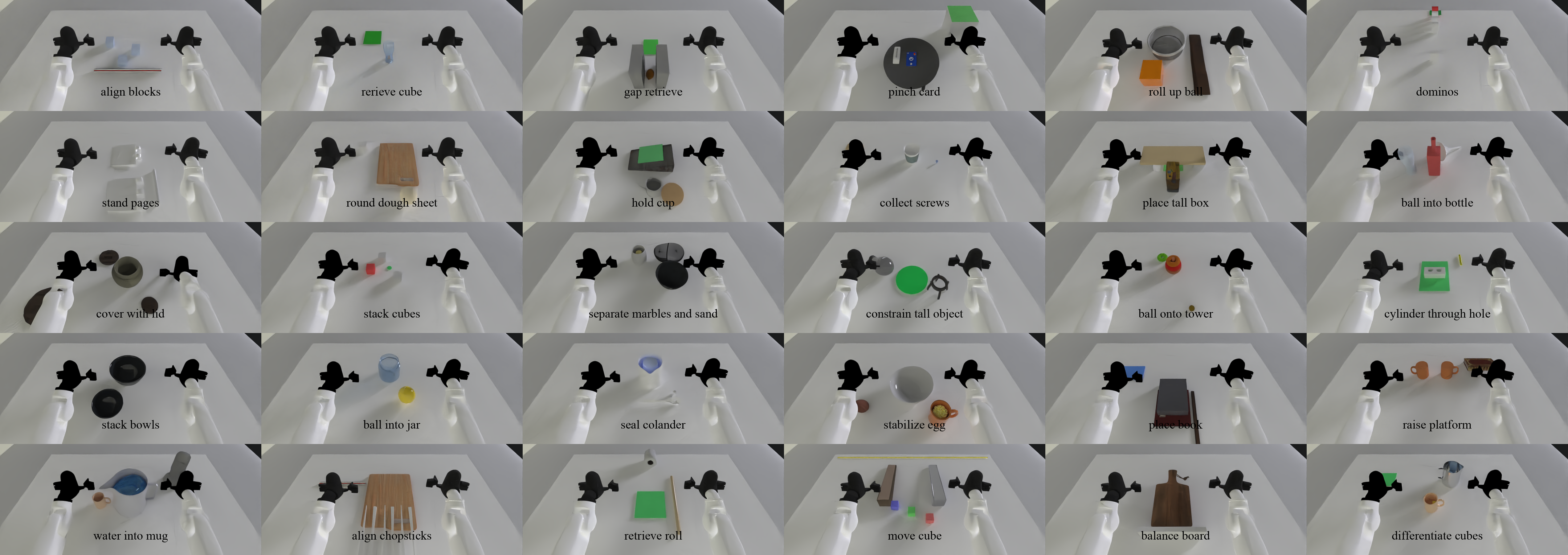}
    \caption{Ego-centric view of the 30 seed tasks in \papername.}
    \label{fig:gallery_ego}
\end{figure*}

The details of 30 seed tasks are as follows. The ego-centric view of the seed task gallery is in Figure~\ref{fig:gallery_ego}.

\begin{enumerate}
\item[\textbf{01}] \textbf{align blocks}: Align the three cubes perfectly in a straight line.
  \\ The task is to arrange multiple scattered cubes into a precise linear sequence. Attempting this cube-by-cube with a standard gripper is inefficient and prone to misalignment. The solution exploits the \textbf{one-dimensional geometric property} of the ruler's edge to act as a physical constraint, ensuring that all objects sharing contact with that edge are perfectly aligned.

\item[\textbf{02}] \textbf{retrieve cube}: Retrieve the cube from the container and place it on the green target area.
  \\ The challenge involves retrieving a target object that is not directly reachable by the robot's end-effector due to geometric constraints. The task requires the robot to first pour the cube out of the deep narrow container and then move it to the target area.

\item[\textbf{03}] \textbf{gap retrieve}: Place the lemon on the green target area.
  \\ The task is to place a lemon on a target area. The lemon is located in a narrow channel. The channel is formed by the left and right boundary blocks, creating a space too narrow for a standard robotic gripper to enter. The solution requires first moving one boundary block to form a larger gap, then grasping the lemon and placing the lemon on the target area.

\item[\textbf{04}] \textbf{pinch card}: Pick the bank card up from the 'small table' surface.
  \\ The 'bank card' is too thin and slippery for a standard gripper to push directly on a flat surface. The robot must utilize the friction and deformability of the 'eraser' to create enough lateral force to move the card. This material property is essential because the high-friction interface allows the robot to translate the card to a position (the small table's edge) where its geometric constraints no longer prevent a successful grasp.

\item[\textbf{05}] \textbf{roll up box}: Move the box into the basket.
  \\ The task is to move a box vertically onto a basket. Since the box is too heavy to lift, the robot must use the inclined plane (a simple machine formed by the board and the basket edge) to reduce the required force. This utilizes the box's sliding property to smoothly ascend the ramp surface.

\item[\textbf{06}] \textbf{dominos}: Push the red block onto the green target area.
  \\ The `red block' and `white block 1,2,3' are out of reachability, but happen to form a domino structure when they fall. The robot must use the `white block 4' within the reachability to reach `white block 3'. To make a perfect chain of falling, the robot must push the upper part of `white block 3' and finally, the `red block' is pushed onto the `target area'.

\item[\textbf{07}] \textbf{stand pages}: Assemble the two pages using the stabilizing bar to create a standing hinge structure.
  \\ Individually, the pages lack the base width to overcome tipping forces. The robot must transform these 2D objects into a stable 3D structure by utilizing a \textbf{pin-joint} mechanism. By inserting the bar through the holes, the robot creates a mechanical constraint that links the two objects, allowing them to lean against each other and support their combined weight against gravity.

\item[\textbf{08}] \textbf{round dough sheet}: Flattern the dough ball and make a perfect round sheet
  \\ The task requires creating a precise round shape. It's difficult to flatten the dough evenly with a parallel gripper, so the robot uses the planar surface of the board to distribute pressure. Achieving a perfect round shape requires utilizing the circular symmetry of the separate cutting-edge object.

\item[\textbf{09}] \textbf{hold cup}: Put the cup on the green target area on the slope without sliding down.
  \\ The 'cup' and 'slope' lack sufficient friction to remain stationary. The robot must utilize the \textbf{surface friction} and \textbf{thin-film flexibility} of the cloth to stabilize the cup. This material property is essential because the coaster acts as a high-friction interface that compensates for the low-friction property of the standard rigid materials (the cup and the table).

\item[\textbf{10}] \textbf{collect screws}: Collect the screw into the container
  \\ The screw is too small to be easily grasped directly. The robot must use the dustpan and spatula to collect the screw first and then put it into the container.

\item[\textbf{11}] \textbf{place tall box}: Move the rectangular box onto the green target area.
  \\ The obstacle enforces a geometric clearance constraint. The beam cannot be moved or raised. The solution relies on changing orientation to reduce the effective height, exploiting the different cross-sectional dimensions of the box to fit under the low beam. Tipping the box to a diagonal lean to and pushing it under the beam is not feasible, because the gripper cannot keep the balance of the box.

\item[\textbf{12}] \textbf{ball into bottle}: Transfer the ball from the blue container into the red bottle.
  \\ The `funnel' provides a wide capture entrance. Gravity guides the ball through the spout into the bottle. The 'ball container' is graspable and can be poured into the funnel opening. Otherwise, without the funnel, directly pouring the ball into the bottle's mouth is very difficult and requires high precision.

\item[\textbf{13}] \textbf{cover with lid}: Cover the pot precisely using one lid.
  \\ This requires reasoning about cross-sectional fit. Only the correctly sized circular lid will cover the opening with minimal excess. The circular geometry and ring reference enable precise centering without measurements.

\item[\textbf{14}] \textbf{stack cubes}: Build a stable two-layer stack so the red cube's center is above the green colored dot.
  \\ A wide base increases stability by lowering the combined center of mass and increasing the support polygon. The geometric arrangement is necessary to keep the top cube centered over the marker.

\item[\textbf{15}] \textbf{separate marbles and sand}: Remove sand from the jar and keep the marbles in the jar.
  \\ Individual grains are too small to pick efficiently. By assembling a colander over a bowl, the robot creates a gravity-powered separator that splits marbles from sand and enables clean transfer into the jar.

\item[\textbf{16}] \textbf{stand bulb}: Stand the tall object with a curved bottom stably within the green target area.
  \\ A tall object with a curved bottom is prone to tipping due to a curved base. A surrounding ring provides lateral support and increases the effective base, using geometry for stability.

\item[\textbf{17}] \textbf{ball onto tower}: Place the ball stably on top of the tower.
  \\ The tower pole alone cannot support the ball as it would roll off. The base can only fit 2.5 rings inside its height, so stacking all three rings causes the top ring to protrude above the pole, creating a hollow cup-like surface where the ball can rest stably in the center.

\item[\textbf{18}] \textbf{insert cylinder}: Insert the yellow rod into one of the round holes in the white block so it stands upright in the hole.
  \\ A circular peg requires coaxial alignment to pass a circular hole. This uses cross-section geometry and orientation matching.

\item[\textbf{19}] \textbf{stack bowls}: Nest the smaller bowl concentrically inside the larger one.
  \\ Concentric alignment leverages circular geometry; matching rims provide a strong geometric signal for proper nesting.

\item[\textbf{20}] \textbf{ball into jar}: Put the foam ball fully into the jar.
  \\ The 'foam ball' is larger than the 'glass jar' mouth when undeformed. The task leverages the elasticity and compressibility of the foam to change its shape and pass through the restriction, which is not possible with a rigid sphere.

\item[\textbf{21}] \textbf{seal colander}: Make the container hold water for a short time.
  \\ The perforations of the perforated container prevent it from holding liquid. By placing the curved holder onto the perforated container, the robot can pour the water from the pitcher into the curved holder onto the perforated container and verify it holds.

\item[\textbf{22}] \textbf{stabilize egg}: Make the tube stand upright in the bowl without falling for a short time.
  \\ The narrow base of the 'tube' makes it unstable. Granular packing of the 'dry sand' around the base provides distributed support and friction, stabilizing the tube without direct clamping by the gripper.

\item[\textbf{23}] \textbf{place book}: Insert the book into the bookshelf.
  \\ The bookshelf is deep and tightly packed, leaving no gap wide enough for the new book, and the books sit too deep inside the shelf to be separated directly with a parallel gripper. The robot must first use the stick as a pusher to split the books on the shelf and open up space, and then place the new book into the gap.

\item[\textbf{24}] \textbf{raise platform}: Create an elevated platform and place the book on it.
  \\ The goal requires supporting a load above the table. Assembling two equal-height supports and a rigid deck creates a stable platform that can hold the book without touching the table.

\item[\textbf{25}] \textbf{water into mug}: Collect water in the mug without moving the pitcher.
  \\ The 'pitcher' is fixed and cannot be tilted to pour water directly. By dropping the 'heavy large object' into the pitcher, the robot exploits Archimedes' principle of water displacement to raise the water level and cause overflow into the 'mug', demonstrating understanding of fluid displacement physics.

\item[\textbf{26}] \textbf{align chopsticks}: Make all six chopsticks point in the same direction, with their small thin ends all facing the same way.
  \\ The task is to align six chopsticks so they all point in the same direction. Initially, the chopsticks lean on a board with mixed orientations, some with the heavy end on the board, others with the thin end. By pushing a ruler to slide the chopsticks further onto the board, the \textbf{asymmetric mass distribution} causes chopsticks of different orientations to behave differently: those with the heavy end on the board will settle flat, while others remain tilted. This physical cue reveals which chopsticks need to be flipped, enabling efficient sorting and alignment.

\item[\textbf{27}] \textbf{retrieve roll}: Use the long rod to retrieve the out-of-reach roll and move it to the green target area.
  \\ The roll is placed beyond the robot's direct reach, making it impossible to grasp directly. By using the rod as a tool extension, the robot can extend its effective reach. Inserting the rod into the hollow part of the roll creates a mechanical coupling through friction and geometric constraint, allowing the robot to lift and transport the roll as a single rigid-body system. This task demonstrates tool use for reach extension and object manipulation through insertion.

\item[\textbf{28}] \textbf{move cube}: Move any of the three cubes to the left of the goal line.
  \\ This task requires the robot to understand and exploit the \textbf{friction properties} of different surfaces. The smooth slope has very low friction, allowing objects to slide freely, while the rough slope has very high friction that would stop the cube. The robot must select the correct slope to enable the cube to gain enough momentum to slide past the goal line. This tests the robot's ability to reason about material properties and their effect on object dynamics.

\item[\textbf{29}] \textbf{balance board}: Place the board onto the wall and keep it balanced.
  \\ The goal requires balancing a board on a narrow support. The key insight is to recognize that the board must be placed with its center of mass directly above the wall to maintain equilibrium.

\item[\textbf{30}] \textbf{differentiate cubes}: Place the wooden cube to the green target area. Keep the metal cube inside the mug.
  \\ The challenge involves differentiating two objects that appear similar but have different material properties. The robot must use physical reasoning to identify which cube is wooden by exploiting the buoyancy property - wood floats in water while metal sinks. This requires the robot to perform an intermediate action (pouring water) to reveal the hidden material property before completing the placement task.

\end{enumerate}

\section{Method Details}

\subsection{Mutation and Verification Process}
\label{sec:mut_and_ver}

The mutation and verification process can be summarized as Algorithm~\ref{alg:mutate_and_verify}.
\begin{algorithm}
  \caption{Task Mutation and Verification Process}
  \label{alg:mutate_and_verify}
  \begin{algorithmic}[1] 
    \State {\bfseries Input:} seed task $T_0$, max mutation steps $N$, max refinement rounds $R{=}3$
    \State Initialize $remainingSteps \gets N$, $strategies \gets$ \{\textit{pivot, trap, add}\}
    \State Initialize $taskPool \gets \{T_0\}$, $stage \gets$ \textit{early} \Comment{prioritize \textit{pivot} early, then mix in \textit{trap}/\textit{add}}
    \State Initialize $applyCount[T, st] \gets 0$ for all $T$ and $st$
    \Repeat
      \State Randomly sample a task $T \in taskPool$
      \State Choose a strategy $st \sim P(strategies|stage)$
      \If{$applyCount[T, st] \ge 2$}
        \State {\bfseries continue} \Comment{avoid repeated use of the same strategy on the same task}
      \EndIf
      \State $T' \gets$ \textsc{get\_mutation}($T, st, \emptyset, \emptyset$)
      \For{$i{=}1$ {\bfseries to} $R$}
        \State $vResult \gets$ \textsc{get\_verification}($T'$)
        \If{$vResult.verified$}
          \State $taskPool \gets taskPool \cup \{T'\}$
          \State $applyCount[T, st] \gets applyCount[T, st] + 1$
          \State {\bfseries break}
        \Else 
          \State $feedback \gets vResult.feedback$
          \State $T' \gets$ \textsc{get\_mutation}($T, st, T', feedback$)
        \EndIf
      \EndFor
      \State $remainingSteps \gets remainingSteps - 1$
      \If{\textsc{should\_switch\_stage}($taskPool$)}
        \State $stage \gets$ \textit{late}
      \EndIf
    \Until{$remainingSteps \le 0$}
    \State {\bfseries Output: } mutated tasks $taskPool$
  \end{algorithmic}
\end{algorithm}

\subsection{An Illustrative Case: \textit{Retrieve Cube}}
\label{sec:case}

As shown in Figure~\ref{fig:mutation_retrive_cube}, we use the \textit{retrieve cube} task as a running example to illustrate how different mutation strategies generate diverse task variants. The seed task \textit{retrieve cube} requires the robot to retrieve a cube from a narrow-opening container and place it onto a green target area (\textit{Initial}). Since the opening is too small for the gripper to enter, a natural potential solution is to lift the container, pour the cube out, and then place the cube onto the target area. 

During the \textit{pivot} phase, one effective way to block the original solution is to fix the narrow-open container to the table, making it impossible to lift and tilt anymore. Under this constraint, one pivot introduces a long spoon with a small head, enabling the robot to scoop the cube out (\textit{Pivot1}). Another pivot adds a cup of water and specifies the cube has low density, allowing it to float upward until it becomes reachable (\textit{Pivot2}). Building on (\textit{Pivot2}), the water cup is replaced with a water tank, making direct pouring infeasible. The task is then pivoted again by adding a kitchen ladle, which allows the robot to transfer water from the tank into the container (\textit{Pivot3}).

As examples of \textit{trap} mutations, based on (\textit{Pivot1}), a large spoon is added as a trap (\textit{Trap1}), since it appears useful for scooping but it can't fit in the narrow opening. Another trap adds a cup of water (\textit{Trap2}) to \textit{Initial}, which may look helpful but leads to a less efficient and more difficult strategy compared to scooping. Also based on (\textit{Initial}), a wooden skewer is added to mislead an implausible strategy of stabbing the cube and pulling it out (\textit{Trap3}); however, the cube is rigid and can't be penetrated, making the skewer inefficient.

Finally, for \textit{add} mutations, tasks sampled from the task pool (\textit{Initial, Pivot1-3, Trap1-3}) are augmented with random distractor objects from a predefined list. In this example, distractors include an IKEA plate, a TV remote controller, a copper kettle, a lemon, a paper cup, a water bottle with handle and pump, and a wrench, which form additional cluttered variants (\textit{Add1-3}).

\subsection{Example Mutation}

\begin{figure*}[h]
    \centering
    \includegraphics[width=\linewidth]{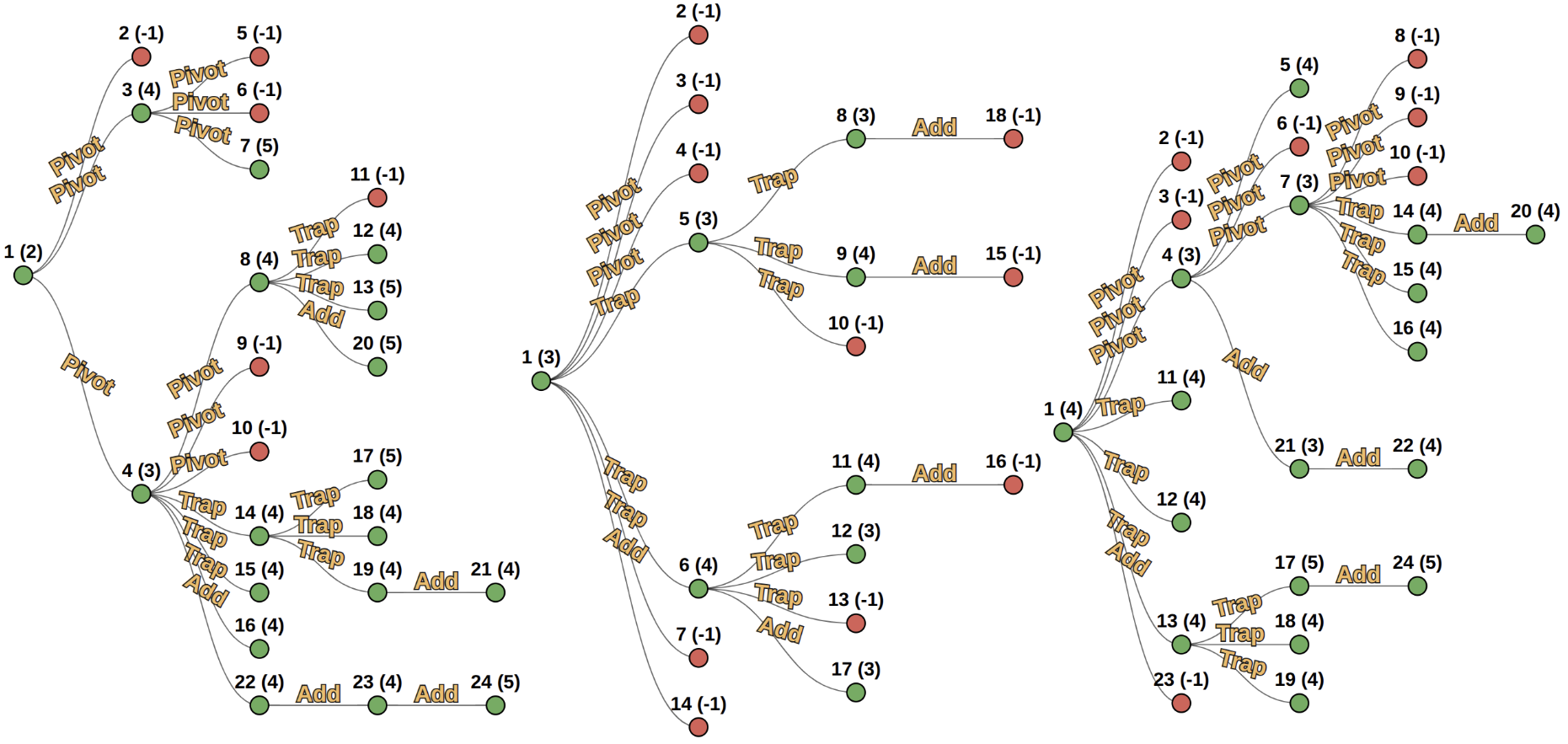}
    \caption{Task mutation trees for \textit{retrieve cube}, \textit{gap retrieve}, and \textit{roll up box}. The root node represents the seed task. Edges show mutation types: \textit{pivot}, \textit{trap}, or \textit{add}. Green nodes indicate successful mutations, while red nodes indicate failures. Node labels X(Y) denote the Mutation ID and task difficulty score, respectively.}
    \label{fig:task_mutation_trees}
\end{figure*}

Fig~\ref{fig:task_mutation_trees} depicts three examples of mutation trees.

\section{Baseline Details}

\texttt{VLM Controller} and \texttt{VLM Planner} are agents based on \texttt{gpt-4o}, who has access to different levels of oracles. The prompt templates can be found in Appendix~\ref{sec:controller_prompt} and ~\ref{sec:planner_prompt}.

\texttt{ACT} is trained for 100,000 steps with a chunk size of 50, batch size of $256$.

\texttt{$\pi_0$} is fine-tuned for 100,000 steps with an effective batch size of
$64$.

\texttt{$\pi_{0.5}$} is fine-tuned for 100,000 steps with an effective batch size of $64$.

\subsection{Compute}

We trained baseline models on 8 RTX 6000 for two days on average.

\section{Additional Benchmark Analysis}
\label{sec:benchmark_analysis}

\subsection{Benchmark Coverage Statistics}
\label{sec:app_coverage}

To better illustrate the coverage of the tasks in our benchmark, we compiled the following statistics.

\begin{table}[h]
\centering
\caption{Coverage statistics of our benchmark tasks.}
\label{tab:benchmark_coverage}
\begin{tabular}{lcl}
\toprule[1.2pt]
 & \textbf{Count} & \textbf{Examples} \\
\midrule
\textbf{Object Instances} & 264 & cup, sponge, ruler, block, ... \\
\textbf{Object Categories} & 13 & containers, tools, food, toys, ... \\
\textbf{Spatial Relations} & 14 & on, inside, between, under, ... \\
\textbf{Manipulation Primitives} & 32 & lift, push, pour, scoop, ... \\
\textbf{Reasoning Patterns} & $>70$ & 30 Seed + 43 Pivot \\
\bottomrule[1.2pt]
\end{tabular}
\end{table}

\subsection{Human Validation of Generated Tasks}
\label{sec:app_human_study}

To check that the generated tasks are solvable, free of shortcuts, and correctly rated for difficulty, we ran a human validation study with three independent annotators. We randomly sampled 60 tasks, 15 each from seed, pivot, trap, and add tasks. Annotators attempted each task by teleoperation in simulation, with a 2-minute limit per trial, and recorded the number of trials needed to succeed or marked the task unsolvable. They then answered a questionnaire on three criteria: whether the task is free of shortcuts, whether the generated intended solution is feasible, and whether they agree with the assigned difficulty score. All 60 sampled tasks were solved by teleoperation. Table~\ref{tab:app_human_study} summarizes the results.

\begin{table}[h]
\centering
\caption{\textbf{Human validation study} on 60 randomly sampled tasks (15 per task type). The first row reports the average number of teleoperation trials needed to solve a task; the remaining rows report the number of tasks (out of 15) that meet each criterion.}
\label{tab:app_human_study}
\begin{tabular}{lccccc}
\toprule[1.2pt]
 & \textbf{Seed} & \textbf{Pivot} & \textbf{Trap} & \textbf{Add} & \textbf{Overall} \\
\midrule
Avg.\ \# trials to solve & 2.3 & 2.5 & 1.9 & 1.5 & -- \\
Shortcut-free & 15 & 14 & 15 & 14 & 96.7\% \\
Feasible generated solution & 15 & 15 & 12 & 13 & 91.7\% \\
Agrees with difficulty score & 14 & 13 & 12 & 14 & 88.3\% \\
\bottomrule[1.2pt]
\end{tabular}
\end{table}

\subsection{Task Difficulty and Model Performance}
\label{sec:app_difficulty}

\begin{figure}[h]
    \centering
    \includegraphics[width=0.9\linewidth]{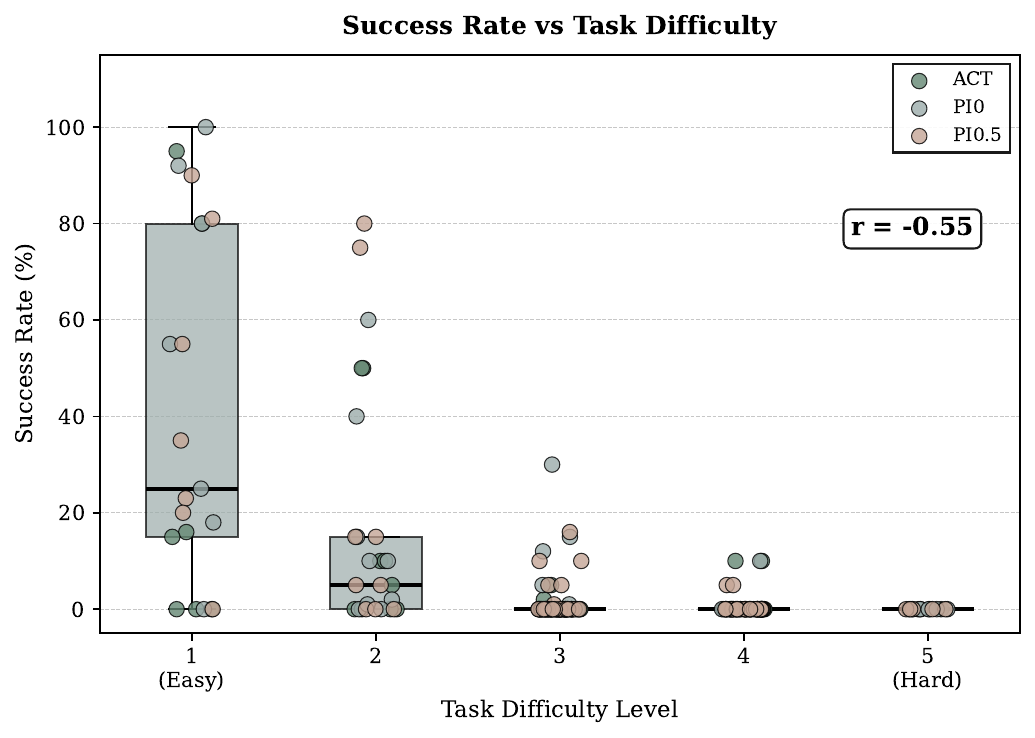}
    \caption{\textbf{Success Rate vs task difficulty.} Box plots show score distributions at each difficulty level with individual model observations overlaid. Pearson correlation coefficient of $r = -0.55$ confirms a moderate negative relationship between difficulty and performance, indicating that our difficulty ratings possess meaningful predictive validity.}
    \label{fig:app_difficulty}
\end{figure}

To check whether the task difficulty scores capture real task complexity, we compare each task's difficulty level with the VLA baselines' measured performance. Success rate falls steadily as difficulty rises (Table~\ref{tab:app_difficulty_sr}, Figure~\ref{fig:app_difficulty}), and the Pearson correlation between difficulty and success is $r = -0.55$.

\begin{table}[h]
\centering
\caption{\textbf{VLA success rate by task difficulty level.}}
\label{tab:app_difficulty_sr}
\begin{tabular}{lccccc}
\toprule[1.2pt]
\textbf{Difficulty level} & 1 & 2 & 3 & 4 & 5 \\
\midrule
\textbf{Success rate} & 47.1\% & 9.2\% & 1.7\% & 0.7\% & 0.0\% \\
\bottomrule[1.2pt]
\end{tabular}
\end{table}

\subsection{Diversity of Human Demonstrations}
\label{sec:app_demo_diversity}

\begin{figure}[h]
    \centering
    \includegraphics[width=1.0\linewidth]{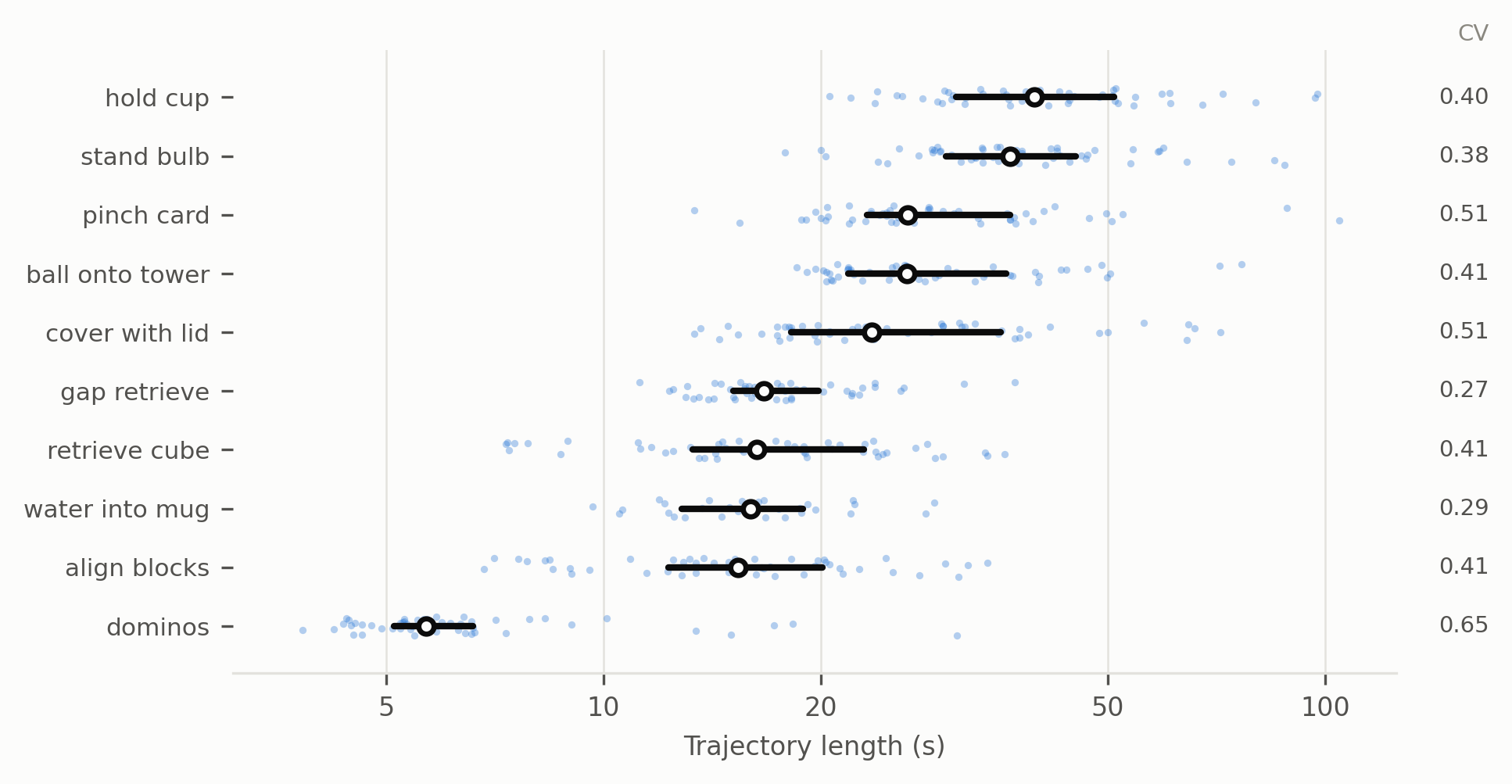}
    \caption{\textbf{Distribution of trajectory length per task.} Each light marker is one demonstration; the black bar spans the interquartile range, and the open circle marks the median. Tasks are ordered by mean length, and the x-axis is logarithmic. The right-hand column reports the coefficient of variation of length within each task; the median across tasks is 0.41.}
    \label{fig:app_traj_length}
\end{figure}

\begin{figure}[h]
    \centering
    \includegraphics[width=1.0\linewidth]{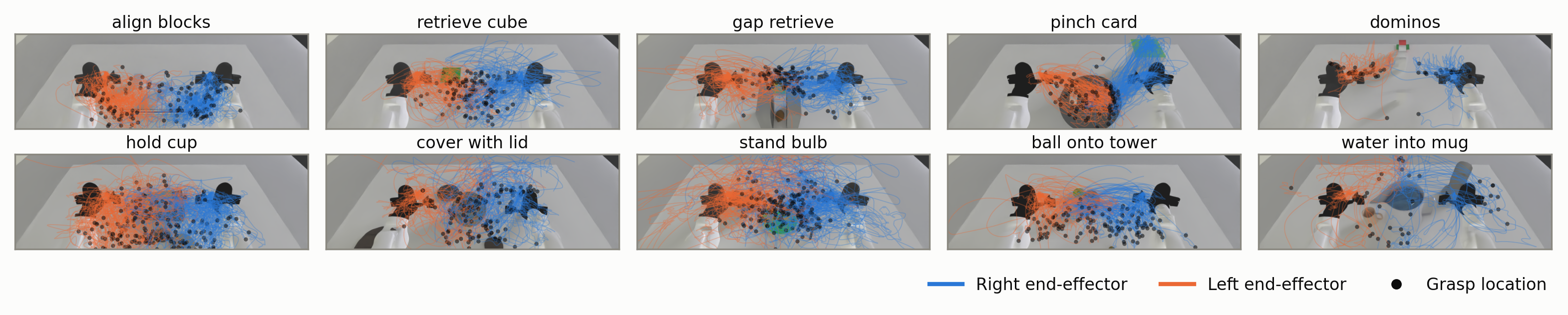}
    \caption{\textbf{End-effector paths of all demonstrations.} Every demonstration's end-effector path is projected into the image plane of the ego camera. Blue and orange traces are the right and left end-effector paths, respectively, one trace per demonstration; black markers are the positions at which the gripper closed. Approach paths and grasp locations occupy broad continuous regions rather than a small number of narrow tubes.}
    \label{fig:app_ee_paths}
\end{figure}

\begin{figure}[h]
    \centering
    \includegraphics[width=0.8\linewidth]{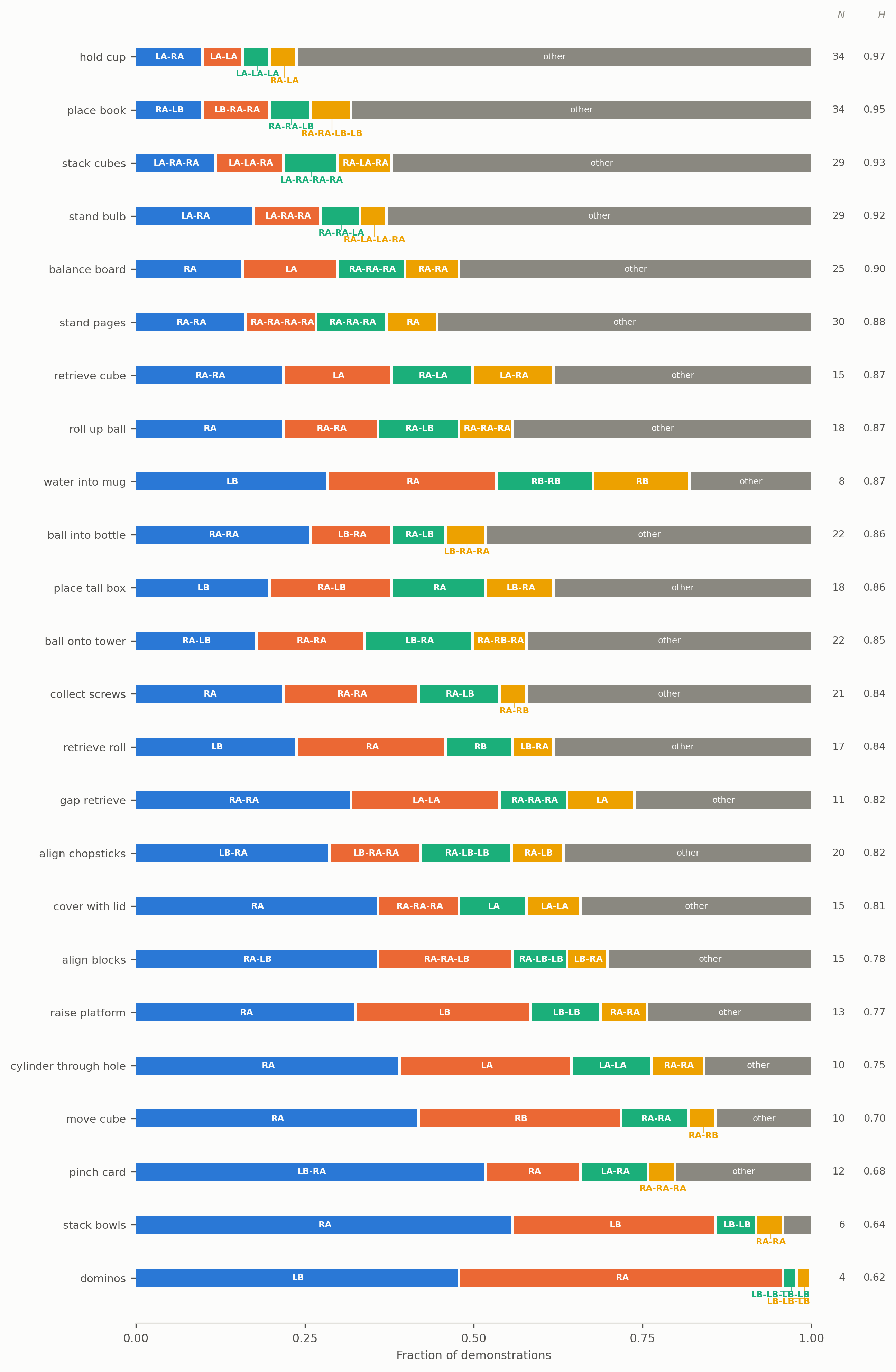}
    \caption{\textbf{Composition of contact orders per task.} Each demonstration is encoded as an ordered sequence of arm+site tokens, where \textit{R}/\textit{L} denote the right and left arm and \textit{A/B/C} denote automatically identified grasp sites; for example, \textit{RA-LB} is a right-arm grasp at site A followed by a left-arm grasp at site B. Bars show the four most frequent orders per task, with the remaining orders pooled in grey. \textit{N} is the total number of distinct orders observed, and \textit{H} is the normalized entropy of their distribution (1.0 = uniform). The most frequent order accounts for at most 52\% of demonstrations in any task.}
    \label{fig:app_contact_orders}
\end{figure}

\begin{figure}[h]
    \centering
    \includegraphics[width=1.0\linewidth]{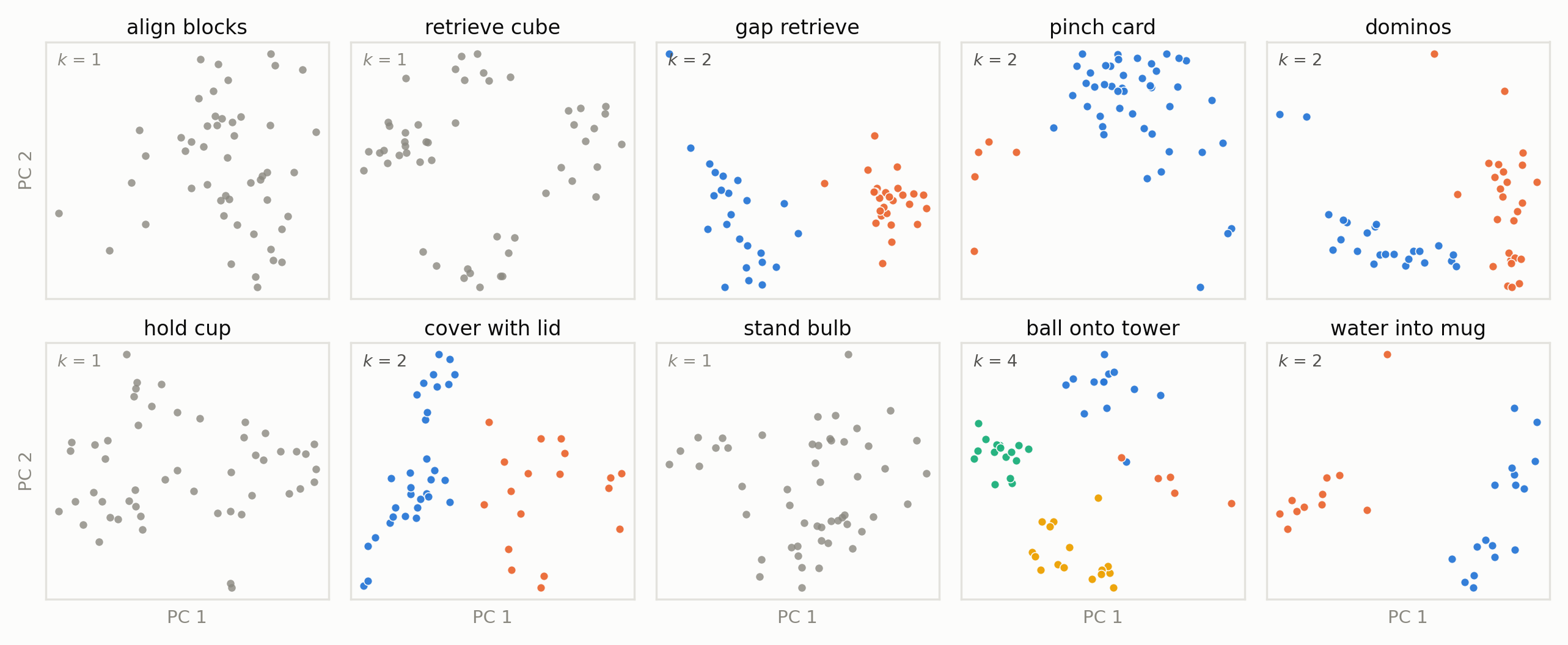}
    \caption{\textbf{Solution modes from trajectory clustering.} Each point is one trajectory, projected onto the first two principal components of its time-normalized 14-D trajectory. Color indicates the k-means cluster assignment; \textit{k} is chosen per task by silhouette score, subject to every cluster containing at least 10\% of demonstrations, and tasks where no such split is supported are shown in grey with \textit{k} = 1. Six of ten tasks admit two or more solution modes.}
    \label{fig:app_solution_modes}
\end{figure}

A low seed-task score could mean the demonstrations follow a single narrow trajectory that the policy merely replays. To check this, we analyze the collected human demonstrations along five axes:
\begin{itemize}[leftmargin=*, itemsep=0pt]
    \item \textbf{Trajectory length} (Figure~\ref{fig:app_traj_length}). The median coefficient of variation of episode duration is 0.41, ranging from 0.27 (\textit{Gap Retrieve}) to 0.65 (\textit{Dominos}). Within a task, the longest demonstration is up to about $4.9\times$ as long as the shortest.
    \item \textbf{Action-sequence diversity.} The mean pairwise dynamic time warping (DTW) distance over the 6-D bimanual end-effector positions gives a median normalized DTW of 0.55 across tasks (range 0.39--0.59). Figure~\ref{fig:app_ee_paths} visualizes the end-effector paths.
    \item \textbf{Contact order} (Figure~\ref{fig:app_contact_orders}). We detect grasp events from the commanded gripper width, cluster the grasp sites, and encode each demonstration as an ordered string of arm and grasp-site tokens. At the median, the most frequent order covers 30\% of a task's demonstrations, and it never exceeds 52\%. The normalized entropy is 0.84 of its maximum. \textit{Hold Cup} shows 34 distinct contact orders across 50 demonstrations, and its most frequent order appears only 5 times. In \textit{Dominos}, demonstrations split 48\% / 48\% between the left and right arm.
    \item \textbf{Solution modes} (Figure~\ref{fig:app_solution_modes}). We resample each trajectory to a fixed length in the 14-D bimanual pose-and-gripper space, then apply PCA and k-means, requiring every cluster to hold at least 10\% of the demonstrations. Six of ten tasks admit two or more modes, and \textit{Ball Tower} admits four. Silhouette scores are modest (0.27--0.39), which indicates that within-task variation is largely a continuum. Discrete multimodality is concentrated in contact order.
    \item \textbf{Initial scene variation} (Section~\ref{sec:app_scene_variation}).
\end{itemize}

Together, these analyses show that the demonstrations cover a broad range of behaviors rather than a single scripted solution.

\subsection{Skill Coverage of Seed Demonstrations}
\label{sec:app_skill_coverage}

We manually extracted 18 core manipulation primitives from the 30 seed tasks: \textit{push down}, \textit{push forward}, \textit{pick}, \textit{place}, \textit{pour}, \textit{press}, \textit{lift}, \textit{grasp}, \textit{open}, \textit{sweep}, \textit{dump}, \textit{insert}, \textit{stack}, \textit{pull}, \textit{set aside}, \textit{move}, \textit{tilt}, and \textit{slide}. Most skills required by the mutation tasks already appear in the seed demonstrations. One clear exception is \textit{scoop}, which the \textit{Water Mug} pivot mutation requires. We deliberately do not provide a complete predefined skill set, because it would reveal the intended solutions. Overcoming a missing skill is part of the creative problem solving that \papername is designed to evaluate.

\subsection{Scene Variation between Training and Evaluation}
\label{sec:app_scene_variation}

\begin{figure}[h]
    \centering
    \includegraphics[width=1.0\linewidth]{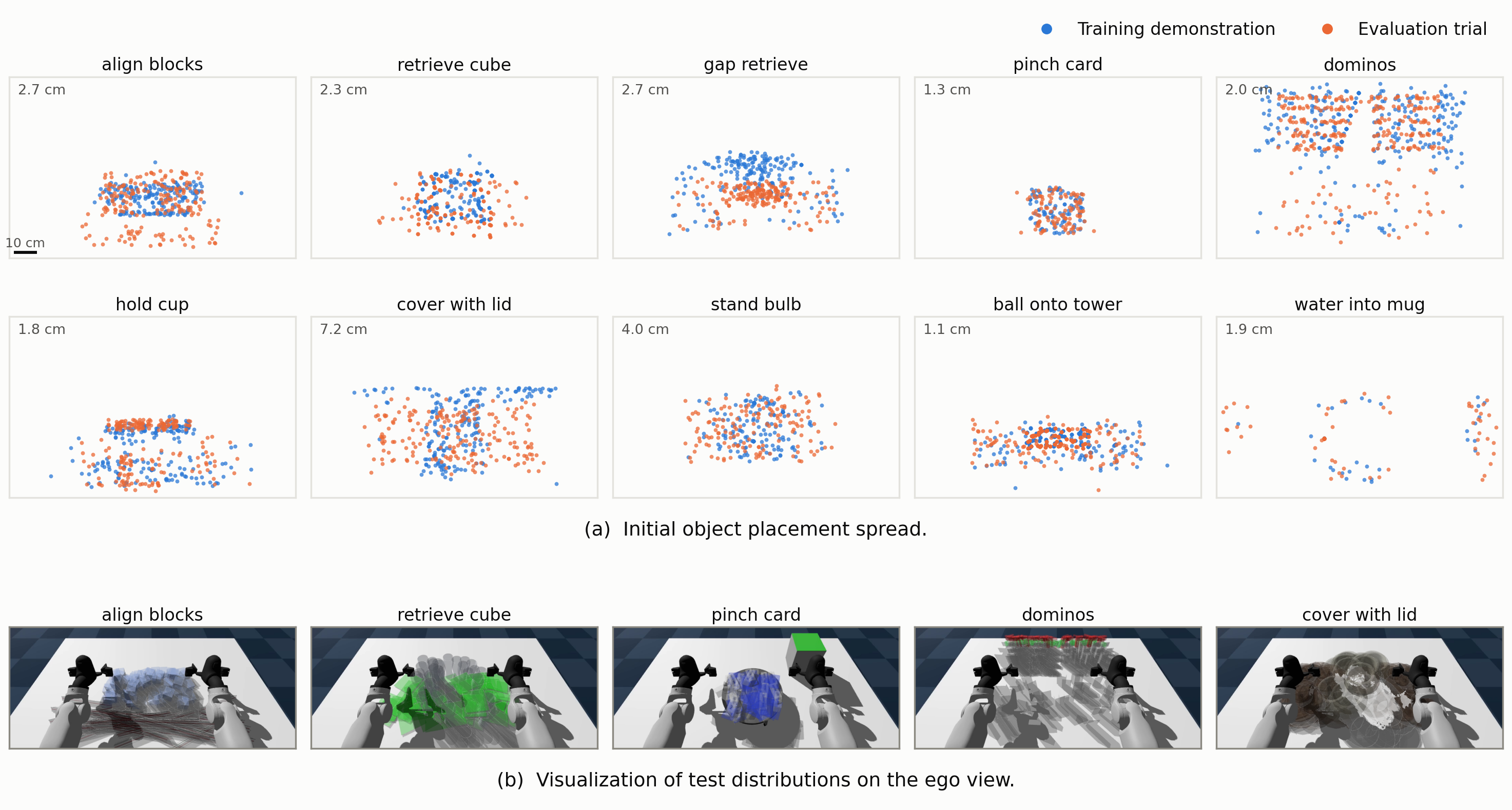}
    \caption{(a) \textbf{Initial object placement spread in a top-down view.} Each marker is one randomized object in one episode, and the number in each panel is the median distance from an evaluation placement to the nearest training placement of the same object. (b) \textbf{Visualization of test distributions on the ego view.} All evaluation placements for five representative tasks, overlaid on the ego camera image.}
    \label{fig:app_scene_variation}
\end{figure}

Evaluation episodes use object placements that are randomized separately from the training demonstrations. Figure~\ref{fig:app_scene_variation} shows how widely objects are spread at initialization and how far each evaluation placement lies from the nearest training placement. Table~\ref{tab:app_spatial_variance} compares the spread of placements in training and evaluation across all 37 randomized objects. A typical object covers a region of about $19\,\text{cm} \times 43\,\text{cm}$. The median distance from an evaluation placement to the nearest training placement of the same object is 4.1\,cm, and 4\% of evaluation placements fall outside the training bounds on at least one axis.

\begin{table}[h]
\centering
\caption{\textbf{Spread of initial object placements} in training and evaluation episodes (median over 37 randomized objects).}
\label{tab:app_spatial_variance}
\begin{tabular}{lcc}
\toprule[1.2pt]
 & \textbf{Training} & \textbf{Evaluation} \\
\midrule
Position std.\ dev.\ ($x$) & 5.4\,cm & 5.6\,cm \\
Position std.\ dev.\ ($y$) & 12.4\,cm & 15.3\,cm \\
Yaw range & 58$^\circ$ & 59$^\circ$ \\
\bottomrule[1.2pt]
\end{tabular}
\end{table}

\section{Additional Experiments and Results}
\label{sec:additional_results}

\subsection{VLM Baselines with Gemini 3 Pro}
\label{sec:app_gemini}

We conducted additional evaluations of the VLM Planner using Gemini 3 Pro. As shown in Table~\ref{tab:gemini_eval}, the overall performance is comparable to GPT-4o, although we observed slightly higher API latency with Gemini.

\begin{table}[h]
\centering
\caption{Comparison of VLM Planner performance using GPT-4o and Gemini 3 Pro.}
\label{tab:gemini_eval}
\begin{tabular}{lcc}
\toprule[1.2pt]
 & \textbf{GPT-4o} & \textbf{Gemini 3 Pro} \\
\midrule
\textbf{Seed SR} & 60.8\% & 69.6\% \\
\textbf{Mutation SR} & 44.4\% & 47.3\% \\
\bottomrule[1.2pt]
\end{tabular}
\end{table}

For the results reported in the paper, we keep a minimalist prompt design, focusing on standardizing the I/O format and defining available action primitives. We tried prompt tuning and did not observe obvious differences. From our observation, the planner is not sensitive to specific prompt engineering, and the bottleneck lies in the model's core reasoning ability.

\subsection{Repeated Evaluation with Standard Errors}
\label{sec:app_repeated_eval}

To measure how much the main results vary between runs, we repeat the evaluation of all baselines on the 10 seed tasks and their mutations three times, each run with a different random seed. We also raise the number of trials per mutation task from 5 to 20. Tables~\ref{tab:app_repeated_sr} and~\ref{tab:app_repeated_ps} report the mean and standard error over the three runs. These are separate runs, so individual numbers differ slightly from Table~\ref{tab:main}. The conclusions hold: every VLA drops from seed to mutation tasks, for example $\pi_0$ falls from $11.9 \pm 0.6\%$ to $7.2 \pm 0.7\%$ average SR and $\pi_{0.5}$ from $12.3 \pm 0.3\%$ to $8.1 \pm 1.5\%$. The oracle-state planners have large standard errors on some tasks (e.g., \textit{Stand Bulb} for \texttt{VLM Planner$^{**}$}), so we treat their per-task numbers as indicative only.

\begin{table}[h]
\centering
\caption{\textbf{Success rate (\%) over three independent evaluation runs}, reported as mean$_{\pm\text{SE}}$. Mutation tasks are evaluated with 20 trials each. * Models assume oracle state. ** Models assume perfect action primitives.}
\label{tab:app_repeated_sr}
\resizebox{\textwidth}{!}{%
\begin{tabular}{llccccccccccc}
\toprule
\textbf{Method} & \textbf{Type} & \textbf{Align Blocks} & \textbf{Retr. Cube} & \textbf{Dominos} & \textbf{Gap Retr.} & \textbf{Stand Bulb} & \textbf{Cover Lid} & \textbf{Ball Tower} & \textbf{Hold Cup} & \textbf{Pinch Card} & \textbf{Water Mug} & \textbf{Avg.} \\
\midrule
ACT & Seed & $16.0_{\pm 4.0}$ & $1.3_{\pm 1.2}$ & $80.7_{\pm 2.3}$ & $3.3_{\pm 2.3}$ & $0.7_{\pm 1.2}$ & $0.0_{\pm 0.0}$ & $0.0_{\pm 0.0}$ & $0.0_{\pm 0.0}$ & $0.7_{\pm 1.2}$ & $0.0_{\pm 0.0}$ & $10.3_{\pm 0.7}$ \\
 & Mut. & $5.7_{\pm 1.2}$ & $0.0_{\pm 0.0}$ & $34.3_{\pm 1.2}$ & $5.3_{\pm 3.1}$ & $0.0_{\pm 0.0}$ & $0.0_{\pm 0.0}$ & $0.0_{\pm 0.0}$ & $0.0_{\pm 0.0}$ & $0.0_{\pm 0.0}$ & $0.0_{\pm 0.0}$ & $4.5_{\pm 0.3}$ \\
\midrule
$\pi_0$ & Seed & $16.7_{\pm 2.3}$ & $1.3_{\pm 1.2}$ & $86.0_{\pm 5.3}$ & $13.3_{\pm 3.1}$ & $0.0_{\pm 0.0}$ & $0.0_{\pm 0.0}$ & $0.0_{\pm 0.0}$ & $0.7_{\pm 1.2}$ & $0.7_{\pm 1.2}$ & $0.0_{\pm 0.0}$ & $11.9_{\pm 0.6}$ \\
 & Mut. & $12.0_{\pm 1.7}$ & $0.0_{\pm 0.0}$ & $55.7_{\pm 7.5}$ & $4.0_{\pm 4.0}$ & $0.0_{\pm 0.0}$ & $0.7_{\pm 1.2}$ & $0.0_{\pm 0.0}$ & $0.0_{\pm 0.0}$ & $0.0_{\pm 0.0}$ & $0.0_{\pm 0.0}$ & $7.2_{\pm 0.7}$ \\
\midrule
$\pi_{0.5}$ & Seed & $23.3_{\pm 1.2}$ & $0.7_{\pm 1.2}$ & $80.7_{\pm 1.2}$ & $16.7_{\pm 1.2}$ & $0.0_{\pm 0.0}$ & $0.0_{\pm 0.0}$ & $0.0_{\pm 0.0}$ & $0.7_{\pm 1.2}$ & $0.7_{\pm 1.2}$ & $0.0_{\pm 0.0}$ & $12.3_{\pm 0.3}$ \\
 & Mut. & $11.3_{\pm 4.2}$ & $1.3_{\pm 2.3}$ & $58.3_{\pm 4.2}$ & $4.0_{\pm 3.5}$ & $0.0_{\pm 0.0}$ & $1.3_{\pm 1.2}$ & $1.7_{\pm 2.9}$ & $3.3_{\pm 4.2}$ & $0.0_{\pm 0.0}$ & $0.0_{\pm 0.0}$ & $8.1_{\pm 1.5}$ \\
\midrule
VLM Controller$^{*}$ & Seed & $17.5_{\pm 24.7}$ & $10.0_{\pm 14.1}$ & $10.0_{\pm 14.1}$ & $2.5_{\pm 3.5}$ & $15.0_{\pm 21.2}$ & $0.0_{\pm 0.0}$ & $5.0_{\pm 7.1}$ & $5.0_{\pm 7.1}$ & $0.0_{\pm 0.0}$ & $0.0_{\pm 0.0}$ & $9.3_{\pm 3.2}$ \\
 & Mut. & $28.2_{\pm 7.3}$ & $2.0_{\pm 2.8}$ & $0.0_{\pm 0.0}$ & $3.0_{\pm 4.2}$ & $7.2_{\pm 10.1}$ & $0.0_{\pm 0.0}$ & $2.0_{\pm 2.8}$ & $0.0_{\pm 0.0}$ & $0.0_{\pm 0.0}$ & $0.0_{\pm 0.0}$ & $8.1_{\pm 4.7}$ \\
\midrule
VLM Planner$^{**}$ & Seed & $56.0_{\pm 5.7}$ & $97.0_{\pm 4.2}$ & $0.0_{\pm 0.0}$ & $80.0_{\pm 0.0}$ & $44.0_{\pm 48.1}$ & $87.5_{\pm 10.6}$ & $100.0_{\pm 0.0}$ & $97.5_{\pm 3.5}$ & $12.0_{\pm 17.0}$ & $0.0_{\pm 0.0}$ & $57.4_{\pm 4.8}$ \\
 & Mut. & $32.7_{\pm 0.9}$ & $38.0_{\pm 25.5}$ & $12.5_{\pm 17.7}$ & $56.0_{\pm 33.9}$ & $22.0_{\pm 29.6}$ & $63.0_{\pm 24.0}$ & $86.8_{\pm 1.1}$ & $98.0_{\pm 2.8}$ & $6.0_{\pm 8.5}$ & $1.5_{\pm 2.1}$ & $41.7_{\pm 3.9}$ \\
\bottomrule
\end{tabular}
}
\end{table}

\begin{table}[h]
\centering
\caption{\textbf{Progress score over three independent evaluation runs}, reported as mean$_{\pm\text{SE}}$. \texttt{VLM Planner$^{**}$} assumes perfect action primitives and has no progress score. * Models assume oracle state.}
\label{tab:app_repeated_ps}
\resizebox{\textwidth}{!}{%
\begin{tabular}{llccccccccccc}
\toprule
\textbf{Method} & \textbf{Type} & \textbf{Align Blocks} & \textbf{Retr. Cube} & \textbf{Dominos} & \textbf{Gap Retr.} & \textbf{Stand Bulb} & \textbf{Cover Lid} & \textbf{Ball Tower} & \textbf{Hold Cup} & \textbf{Pinch Card} & \textbf{Water Mug} & \textbf{Avg.} \\
\midrule
ACT & Seed & $0.51_{\pm 0.18}$ & $0.27_{\pm 0.03}$ & $0.95_{\pm 0.01}$ & $0.51_{\pm 0.03}$ & $0.54_{\pm 0.01}$ & $0.48_{\pm 0.02}$ & $0.58_{\pm 0.02}$ & $0.54_{\pm 0.05}$ & $0.50_{\pm 0.09}$ & $0.30_{\pm 0.04}$ & $0.52_{\pm 0.02}$ \\
 & Mut. & $0.41_{\pm 0.20}$ & $0.23_{\pm 0.01}$ & $0.74_{\pm 0.13}$ & $0.49_{\pm 0.01}$ & $0.28_{\pm 0.14}$ & $0.42_{\pm 0.02}$ & $0.51_{\pm 0.01}$ & $0.43_{\pm 0.01}$ & $0.44_{\pm 0.06}$ & $0.27_{\pm 0.01}$ & $0.42_{\pm 0.04}$ \\
\midrule
$\pi_0$ & Seed & $0.52_{\pm 0.20}$ & $0.31_{\pm 0.02}$ & $0.97_{\pm 0.01}$ & $0.62_{\pm 0.02}$ & $0.53_{\pm 0.02}$ & $0.54_{\pm 0.02}$ & $0.62_{\pm 0.04}$ & $0.62_{\pm 0.05}$ & $0.62_{\pm 0.02}$ & $0.37_{\pm 0.06}$ & $0.57_{\pm 0.03}$ \\
 & Mut. & $0.46_{\pm 0.21}$ & $0.23_{\pm 0.02}$ & $0.79_{\pm 0.16}$ & $0.48_{\pm 0.03}$ & $0.29_{\pm 0.15}$ & $0.44_{\pm 0.01}$ & $0.56_{\pm 0.01}$ & $0.45_{\pm 0.04}$ & $0.52_{\pm 0.01}$ & $0.25_{\pm 0.01}$ & $0.45_{\pm 0.05}$ \\
\midrule
$\pi_{0.5}$ & Seed & $0.55_{\pm 0.14}$ & $0.29_{\pm 0.03}$ & $0.96_{\pm 0.00}$ & $0.63_{\pm 0.01}$ & $0.55_{\pm 0.02}$ & $0.55_{\pm 0.02}$ & $0.54_{\pm 0.01}$ & $0.58_{\pm 0.05}$ & $0.51_{\pm 0.02}$ & $0.35_{\pm 0.08}$ & $0.55_{\pm 0.02}$ \\
 & Mut. & $0.47_{\pm 0.15}$ & $0.26_{\pm 0.07}$ & $0.78_{\pm 0.15}$ & $0.49_{\pm 0.03}$ & $0.29_{\pm 0.15}$ & $0.43_{\pm 0.02}$ & $0.52_{\pm 0.03}$ & $0.45_{\pm 0.04}$ & $0.43_{\pm 0.03}$ & $0.22_{\pm 0.03}$ & $0.43_{\pm 0.04}$ \\
\midrule
VLM Controller$^{*}$ & Seed & $0.37_{\pm 0.52}$ & $0.58_{\pm 0.18}$ & $0.67_{\pm 0.08}$ & $0.59_{\pm 0.08}$ & $0.57_{\pm 0.01}$ & $0.44_{\pm 0.09}$ & $0.48_{\pm 0.18}$ & $0.50_{\pm 0.11}$ & $0.56_{\pm 0.01}$ & $0.29_{\pm 0.00}$ & $0.52_{\pm 0.09}$ \\
 & Mut. & $0.57_{\pm 0.09}$ & $0.37_{\pm 0.01}$ & $0.48_{\pm 0.22}$ & $0.52_{\pm 0.35}$ & $0.13_{\pm 0.01}$ & $0.47_{\pm 0.08}$ & $0.45_{\pm 0.16}$ & $0.50_{\pm 0.01}$ & $0.57_{\pm 0.02}$ & $0.27_{\pm 0.01}$ & $0.41_{\pm 0.06}$ \\
\bottomrule
\end{tabular}
}
\end{table}

\subsection{Results on the Full Task Suite}
\label{sec:app_full_suite}

Table~\ref{tab:main} reports results on 10 representative seed tasks. Table~\ref{tab:app_full_suite} extends the evaluation to all 30 seed tasks and their mutations, under the same protocol: 50 trials for each seed task and 20 trials for each mutation task. The seed-to-mutation gap observed on the 10-task subset holds across the full suite.

\begin{table}[h]
\centering
\caption{\textbf{Results on all 30 Seed Tasks and their Mutations on \papername.} Each cell reports Success Rate (SR) and averaged Progress Score (PS). Mutation results are averaged over all mutations of a seed task. Best performance per task and task type is \textbf{bolded}.}
\label{tab:app_full_suite}
\resizebox{\textwidth}{!}{%
\begin{tabular}{lcccccccccccc}
\toprule
& \multicolumn{4}{c}{\textbf{ACT}} & \multicolumn{4}{c}{\boldmath$\pi_0$} & \multicolumn{4}{c}{\boldmath$\pi_{0.5}$} \\
\cmidrule(lr){2-5} \cmidrule(lr){6-9} \cmidrule(lr){10-13}
& \multicolumn{2}{c}{Seed} & \multicolumn{2}{c}{Mut.} & \multicolumn{2}{c}{Seed} & \multicolumn{2}{c}{Mut.} & \multicolumn{2}{c}{Seed} & \multicolumn{2}{c}{Mut.} \\
\cmidrule(lr){2-3} \cmidrule(lr){4-5} \cmidrule(lr){6-7} \cmidrule(lr){8-9} \cmidrule(lr){10-11} \cmidrule(lr){12-13}
\textbf{Task} & SR & PS & SR & PS & SR & PS & SR & PS & SR & PS & SR & PS \\
\midrule
01 Align Blocks & 14.0\% & 0.61 & 10.0\% & 0.58 & 14.0\% & \textbf{0.63} & \textbf{20.8\%} & 0.62 & \textbf{20.0\%} & \textbf{0.63} & 15.0\% & \textbf{0.63} \\
02 Retrieve Cube & 2.0\% & 0.29 & 0.0\% & \textbf{0.32} & \textbf{6.0\%} & \textbf{0.30} & 0.8\% & 0.31 & 0.0\% & 0.28 & \textbf{1.7\%} & \textbf{0.32} \\
03 Gap Retrieve & 2.0\% & 0.51 & 0.0\% & 0.43 & 18.0\% & 0.65 & 5.0\% & 0.46 & \textbf{20.0\%} & \textbf{0.66} & \textbf{6.0\%} & \textbf{0.49} \\
04 Pinch Card & 0.0\% & 0.41 & 0.0\% & 0.53 & \textbf{10.0\%} & \textbf{0.63} & \textbf{9.0\%} & \textbf{0.62} & 2.0\% & 0.54 & 7.0\% & 0.59 \\
05 Roll Up Box & 0.0\% & 0.36 & 0.0\% & 0.25 & 0.0\% & 0.42 & 0.0\% & \textbf{0.31} & 0.0\% & \textbf{0.50} & 0.0\% & \textbf{0.31} \\
06 Dominos & 52.0\% & 0.87 & 38.3\% & 0.86 & \textbf{86.0\%} & \textbf{0.97} & \textbf{57.5\%} & \textbf{0.87} & 74.0\% & 0.95 & 52.5\% & 0.84 \\
07 Stand Pages & 0.0\% & 0.24 & 0.0\% & 0.16 & 0.0\% & \textbf{0.29} & 0.0\% & \textbf{0.29} & 0.0\% & \textbf{0.29} & 0.0\% & 0.13 \\
08 Round Dough Sheet & 0.0\% & 0.52 & 0.0\% & 0.52 & 0.0\% & \textbf{0.58} & 0.0\% & \textbf{0.58} & 0.0\% & 0.50 & 0.0\% & 0.52 \\
09 Hold Cup & 0.0\% & 0.53 & 0.0\% & 0.39 & 0.0\% & \textbf{0.66} & 0.0\% & 0.45 & \textbf{2.0\%} & 0.64 & 0.0\% & \textbf{0.46} \\
10 Collect Screws & 0.0\% & 0.26 & 0.0\% & 0.31 & 0.0\% & \textbf{0.29} & 0.0\% & 0.32 & 0.0\% & 0.26 & 0.0\% & \textbf{0.33} \\
11 Place Tall Box & 0.0\% & 0.61 & \textbf{3.0\%} & 0.68 & 0.0\% & 0.79 & 1.0\% & 0.69 & \textbf{6.0\%} & \textbf{0.83} & \textbf{3.0\%} & \textbf{0.70} \\
12 Ball into Bottle & 0.0\% & 0.44 & 0.0\% & 0.32 & 0.0\% & \textbf{0.49} & 0.0\% & \textbf{0.37} & 0.0\% & \textbf{0.49} & 0.0\% & 0.35 \\
13 Cover with Lid & 0.0\% & 0.48 & 0.0\% & 0.37 & 0.0\% & 0.53 & 0.0\% & 0.43 & 0.0\% & \textbf{0.58} & 0.0\% & \textbf{0.48} \\
14 Stack Cubes & 0.0\% & 0.36 & 0.0\% & 0.29 & 0.0\% & \textbf{0.40} & 0.0\% & \textbf{0.31} & 0.0\% & 0.37 & 0.0\% & 0.30 \\
15 Separate Marbles and Sand & 16.0\% & 0.43 & 11.2\% & 0.28 & 36.0\% & \textbf{0.81} & \textbf{30.0\%} & \textbf{0.70} & \textbf{42.0\%} & 0.80 & 17.5\% & 0.53 \\
16 Stand Bulb & 0.0\% & 0.54 & 0.0\% & 0.41 & 0.0\% & \textbf{0.55} & 0.0\% & \textbf{0.42} & 0.0\% & \textbf{0.55} & 0.0\% & \textbf{0.42} \\
17 Ball onto Tower & 0.0\% & 0.55 & 0.0\% & 0.48 & 0.0\% & 0.62 & 0.0\% & 0.54 & 0.0\% & \textbf{0.67} & 0.0\% & \textbf{0.58} \\
18 Insert Cylinder & 0.0\% & 0.30 & 0.0\% & 0.26 & \textbf{2.0\%} & \textbf{0.57} & \textbf{0.8\%} & 0.38 & 0.0\% & 0.39 & \textbf{0.8\%} & \textbf{0.41} \\
19 Stack Bowls & 2.0\% & 0.53 & 0.0\% & 0.30 & 24.0\% & 0.68 & \textbf{0.8\%} & \textbf{0.40} & \textbf{26.0\%} & \textbf{0.73} & \textbf{0.8\%} & 0.35 \\
20 Ball into Jar & \textbf{6.0\%} & 0.55 & \textbf{2.0\%} & 0.30 & \textbf{6.0\%} & \textbf{0.59} & \textbf{2.0\%} & 0.29 & \textbf{6.0\%} & 0.55 & \textbf{2.0\%} & \textbf{0.31} \\
21 Seal Colander & 0.0\% & 0.15 & 0.0\% & 0.16 & 0.0\% & \textbf{0.18} & 0.0\% & \textbf{0.18} & 0.0\% & 0.17 & 0.0\% & 0.17 \\
22 Stabilize Egg & 0.0\% & 0.34 & 0.0\% & 0.36 & 0.0\% & \textbf{0.47} & 0.0\% & \textbf{0.42} & 0.0\% & 0.35 & 0.0\% & 0.38 \\
23 Place Book & \textbf{4.0\%} & 0.21 & 3.0\% & 0.25 & 2.0\% & \textbf{0.30} & 3.0\% & 0.40 & 2.0\% & 0.29 & \textbf{26.0\%} & \textbf{0.54} \\
24 Raise Platform & 4.0\% & 0.29 & 0.0\% & 0.29 & \textbf{20.0\%} & \textbf{0.52} & 0.0\% & 0.32 & 18.0\% & 0.49 & 0.0\% & \textbf{0.35} \\
25 Water into Mug & 2.0\% & \textbf{0.36} & \textbf{0.7\%} & 0.43 & 0.0\% & \textbf{0.36} & 0.0\% & 0.45 & \textbf{4.0\%} & 0.35 & 0.0\% & \textbf{0.47} \\
26 Align Chopsticks & 0.0\% & 0.00 & 0.0\% & 0.06 & \textbf{4.0\%} & \textbf{0.06} & 0.0\% & 0.08 & 0.0\% & 0.01 & 0.0\% & \textbf{0.10} \\
27 Retrieve Roll & 0.0\% & 0.20 & 0.0\% & 0.21 & \textbf{2.0\%} & \textbf{0.22} & \textbf{1.4\%} & \textbf{0.22} & 0.0\% & 0.21 & 0.0\% & 0.21 \\
28 Move Cube & 10.0\% & 0.59 & 0.0\% & 0.45 & \textbf{24.0\%} & \textbf{0.69} & 0.0\% & 0.49 & 12.0\% & 0.66 & \textbf{0.7\%} & \textbf{0.50} \\
29 Balance Board & 0.0\% & 0.26 & 0.0\% & 0.34 & \textbf{2.0\%} & \textbf{0.44} & 0.0\% & 0.35 & 0.0\% & 0.35 & 0.0\% & \textbf{0.38} \\
30 Differentiate Cubes & 0.0\% & 0.27 & 0.0\% & 0.34 & 0.0\% & 0.28 & 0.0\% & \textbf{0.40} & 0.0\% & \textbf{0.30} & \textbf{1.0\%} & 0.37 \\
\midrule
\textbf{Average} & 3.8\% & 0.40 & 2.3\% & 0.36 & \textbf{8.5\%} & \textbf{0.50} & 4.4\% & \textbf{0.42} & 7.8\% & 0.48 & \textbf{4.5\%} & \textbf{0.42} \\
\bottomrule
\end{tabular}%
}
\end{table}

\subsection{Failure Analysis across Mutation Types}
\label{sec:app_mutation_type}

To clarify whether failures stem from reasoning, execution, or other factors, we analyzed success rates (SR) across mutation types. Here, \textit{Add(n)} denotes introducing $n$ irrelevant distractor objects during the \textit{Add} mutation.

\begin{table}[h]
\centering
\caption{Success rates across different mutation types.}
\label{tab:mutation_sr}
\begin{tabular}{lccccc}
\toprule[1.2pt]
 & \textbf{Pivot} & \textbf{Trap} & \textbf{Add(1)} & \textbf{Add(2)} & \textbf{Add(3)} \\
\midrule
\textbf{ACT} & 1.7\% & 5.1\% & 13.3\% & 13.5\% & 4.3\% \\
\boldmath$\pi_0$ & 2.0\% & 6.4\% & 13.3\% & 17.6\% & 11.1\% \\
\boldmath$\pi_{0.5}$ & 3.7\% & 5.4\% & 15.0\% & 12.1\% & 7.7\% \\
\bottomrule[1.2pt]
\end{tabular}
\end{table}

Video observations reveal that the specific physical errors directly correlate with the mutation type.

For \textit{Pivot}, which has the lowest SR, models show no strategy transfer. Physically, they may fail by stubbornly attempting to manipulate immovable objects or continuing original trajectories even when objects are accidentally displaced, indicating a true lack of physical reasoning.

For \textit{Trap} and \textit{Add}, SR drops as more objects are introduced, with Add(1/2) SR $>$ Add(3) SR. The physical errors here are primarily driven by distraction, where models may physically freeze due to out-of-distribution visual data or incorrectly reach distractor objects.

\subsection{Behavioral Failure Analysis on Pivot and Trap Mutations}
\label{sec:app_failure_behavior}

To complement the mutation-type breakdown in Table~\ref{tab:mutation_sr}, we manually inspected 100 failed trials of pivot and trap mutations, sampled from six tasks on which the policies perform reasonably on the seed task. Two of the authors annotated each failure using criteria based on observable behavior:
\begin{itemize}[leftmargin=*, itemsep=0pt]
    \item \textbf{Incorrect high-level strategy}: the policy attempts an interaction that is not part of any feasible solution, e.g., repeatedly pushing an immovable object.
    \item \textbf{Missing embodied skill}: the policy interacts with the correct objects but makes no progress.
    \item \textbf{Poor visual grounding}: the policy reaches for or manipulates the wrong object.
    \item \textbf{Low-level execution error}: the intended interaction is correct but fails in execution, e.g., a dropped grasp.
\end{itemize}
These categories describe behavior and do not establish its cause. For example, a policy that pursues an infeasible strategy may do so because it lacks the motor skill the feasible strategy requires. Table~\ref{tab:app_failure_behavior} reports the distribution. The remaining trials (10\% for pivot, 4\% for trap) show ambiguous failure modes that do not fit cleanly into any category.

\begin{table}[h]
\centering
\caption{\textbf{Distribution of observed failure behaviors} over 100 manually annotated failed trials of pivot and trap mutations.}
\label{tab:app_failure_behavior}
\begin{tabular}{lcc}
\toprule[1.2pt]
\textbf{Observed behavior} & \textbf{Pivot} & \textbf{Trap} \\
\midrule
Incorrect high-level strategy & 63\% & 70\% \\
Missing embodied skill & 17\% & 8\% \\
Poor visual grounding & 4\% & 6\% \\
Low-level execution error & 6\% & 12\% \\
\bottomrule[1.2pt]
\end{tabular}
\end{table}

\subsection{Demonstration Scaling}
\label{sec:app_demo_scaling}

To investigate whether more demonstration data could help models perform better on RoboWits, we train $\pi_0$ with 50 and 200 demonstrations, respectively, for two representative tasks.

\begin{table}[h]
\centering
\caption{Effect of increasing the number of demonstrations on two representative tasks.}
\label{tab:demo_scaling}
\begin{tabular}{lll}
\toprule[1.2pt]
\textbf{Number of Demos} & \textbf{Align Blocks} & \textbf{Dominos} \\
\midrule
\textbf{50} & Seed: 44\%, Mut.: 24\% & Seed: 70\%, Mut.: 47\% \\
\textbf{200} & Seed: 54\%, Mut.: 33\% & Seed: 94\%, Mut.: 52\% \\
\bottomrule[1.2pt]
\end{tabular}
\end{table}

\textit{Dominos} relies on a single optimal strategy, i.e., using a pusher to topple the dominos. Scaling the demonstration data provides a clear, consistent signal to resolve the motor-control bottleneck, improving Seed performance from 70\% to 94\%. In contrast, \textit{Align Blocks} requires closed-loop adjustments across multiple valid strategies, such as gathering blocks before aligning them, or aligning two blocks first and then placing the third. This multi-modal nature explains its marginal improvement from data scaling and its greater robustness.

Most importantly, despite improvements on the Seed tasks, performance on the Mutation tasks remains low. This confirms our main conclusion: current VLAs cannot perform robustly when faced with unexpected reasoning-level challenges. The primary bottleneck is indeed poor reasoning rather than control limitations.

\subsection{Demonstration Scaling on Harder Tasks}
\label{sec:app_demo_scaling_ext}

We repeat the demonstration-scaling experiment of Table~\ref{tab:demo_scaling} on four more tasks. Training data contains only successful seed-task demonstrations, so the model never sees a mutated task or a mutation-specific solution. As Table~\ref{tab:app_demo_scaling_ext} shows, 200 demonstrations raise seed success above zero on every task (up to 26\%), while mutation success stays far lower.

\begin{table}[h]
\centering
\caption{\textbf{Effect of increasing the number of demonstrations} for $\pi_0$ on four additional tasks.}
\label{tab:app_demo_scaling_ext}
\begin{tabular}{lcccccccc}
\toprule[1.2pt]
 & \multicolumn{2}{c}{\textbf{Ball Tower}} & \multicolumn{2}{c}{\textbf{Water Mug}} & \multicolumn{2}{c}{\textbf{Stabilize Egg}} & \multicolumn{2}{c}{\textbf{Insert Cylinder}} \\
\cmidrule(lr){2-3} \cmidrule(lr){4-5} \cmidrule(lr){6-7} \cmidrule(lr){8-9}
\textbf{\# Demos} & Seed & Mut. & Seed & Mut. & Seed & Mut. & Seed & Mut. \\
\midrule
50 & 0\% & 0\% & 0\% & 0\% & 0\% & 0\% & 0\% & 0\% \\
200 & 12\% & 5\% & 26\% & 10\% & 8\% & 0\% & 22\% & 8\% \\
\bottomrule[1.2pt]
\end{tabular}
\end{table}

\subsection{Real-World Experiments}
\label{sec:app_real_world}

To check whether the seed-to-mutation gap also appears on real hardware, we collected 200 real-world demonstrations for each of three tasks and fine-tuned and deployed $\pi_0$ on a real robot. The three tasks cover rich contact (\textit{Insert Cylinder}), granular material (\textit{Stabilize Egg}), and fluid (\textit{Water Mug}). For a fair comparison, the simulated $\pi_0$ is also trained with 200 demonstrations. Each seed task is evaluated for 50 trials, and each of its 5--7 mutations for 20 trials. Real-world mutations match their simulated counterparts semantically and physically, with a few asset substitutions for availability and safety: a different water pitcher and a kiwifruit in place of the egg.

As Table~\ref{tab:app_real_world} shows, seed success is roughly 7--9 points higher in the real world than in simulation, which suggests that the simulation, by trading some physical fidelity for efficiency, lowers the achievable success rate. The sharp drop from seed to mutation tasks persists in the real world, consistent with our simulated findings.

\begin{table}[h]
\centering
\caption{\textbf{Real-world vs.\ simulated results} of $\pi_0$ trained with 200 demonstrations.}
\label{tab:app_real_world}
\begin{tabular}{lcccccc}
\toprule[1.2pt]
 & \multicolumn{2}{c}{\textbf{Insert Cylinder}} & \multicolumn{2}{c}{\textbf{Stabilize Egg}} & \multicolumn{2}{c}{\textbf{Water Mug}} \\
\cmidrule(lr){2-3} \cmidrule(lr){4-5} \cmidrule(lr){6-7}
\textbf{Setting} & Seed & Mut. & Seed & Mut. & Seed & Mut. \\
\midrule
Simulation & 22\% & 8\% & 8\% & 0\% & 26\% & 10\% \\
Real world & 30\% & 12\% & 15\% & 3\% & 35\% & 4\% \\
\bottomrule[1.2pt]
\end{tabular}
\end{table}

\section{Schemas and Prompt Templates}

\subsection{Prompt Template for the Seed Task Generator}
\begin{codebox}{markdown}{Seed Task Generator System Prompt}
You are a Robotics Task Designer generating complex, skill-based challenges for a dual-arm robot equipped with parallel grippers (the workspace is a 1.05m x 1.7m table surface, with a bounds of {"x_min": 0, "x_max": 1.05, "y_min": -0.85, "y_max": 0.85, "z_min": 0.8, "z_max": 0.8}). Each task must be returned as a JSON object, and all the tasks must be contained within a single JSON array.

TASK FORMAT:
```json
[
  {
    "task_name": "a concise and meaningful name of the task",
    "object_list": [
      {
        "object_name": "abstract functional name of the object (e.g., 'straightedge tool' instead of 'ruler', 'high-friction pad' instead of 'rubber mat')",
        "appearance_attribute": [
          "an array of required appearance attributes of the object, including its shape, size, color, high or short, fat or slim, ..."
        ],
        "functional_attribute": [
          "an array of required functional attributes of the object, including its material, density, friction, movability, ..."
        ],
        "potential_instances": [
          "at least 3 daily object names to fit the object with attributes required",
          "...",
          "..."
        ],
        "use_primitive": null,
        "asset_id": null
      }
      // ... continue for all necessary objects
    ],
    "initial_scene_setup": "the initial scene setup to form the task, describe the required position and/or relationship of all objects in the object_list", 
    "task_instruction": "the goal of the task without hinting at the solution",
    "task_success_criteria": "the signals to show the task is completed, easy to quantalize, no subjectives,",
    "potential_solution": "how the task should be solved in the easiest way with the current objects provided",
    "task_description": "describe the task in detail, including the goal, the challenge, and why the potential_solution is valid and efficient for the task."
  }
  // ... continue for more unique tasks
]
```

CONSTRAINTS FOR OBJECTS:
 - DO NOT include articulated objects in the object_list. 
 - DO NOT include objects involving aerodynamics, magnetism, thermodynamics, sticky materials or unintended flaws (e.g., unintended holes, scratches, wear, or micro-grooves).
 - The object must be a very common daily item (e.g., mug, hammer, stapler, apple) for which 3D assets are easily found online.
 - DO NOT include the words 'a', 'an', or 'the' in the 'object_name' of the object.
 - Use ABSTRACT FUNCTIONAL names for 'object_name' that describe the object's role in the task, not specific instances. For example: use "straightedge tool" instead of "ruler", "high-friction pad" instead of "rubber mat", "narrow opening container" instead of "bottle", "flat retrieval target" instead of "bank card". List specific instances in 'potential_instances'.
 - An object in the 'object_list' represents a single object (except for liquid and granularity). If there are multiple objects of the same category, write them separately in the 'object_list' (e.g., "boundary block 1", "boundary block 2").
 - DONOT use mechanism names like "inclined plane" or "gap" in the object_list. Instead, DO list the physical objects that create these mechanisms (e.g., a "support block" and a "ramp board" forming the inclined plane).
 - An object is of a single material type. Instead of adding "a cup of water", add a "cup" and a "water" separately and specify the 'water' fills more than half of the 'cup' in the 'initial_scene_setup'.
 - Only list necessary attributes of the object that is important for the task, not all the appearance or functional attributes. Consider specify attributes to realize ideal constraints, like "out of reachability" to avoid direct manipulation.
 - The 'use_primitive' and 'asset_id' of objects will always be null.

CONSTRAINTS FOR OTHERS:
 - DONOT generate tasks that require breaking, permanently deforming (except for soft, intended-to-be-deformed objects like "dough"), or damaging any shape.
 - Ensure all the objects mentioned in the 'initial_scene_setup', 'task_instruction', 'task_success_criteria' and 'potential_solution' are in the 'object_list', except the table and the robot. 
 - Ensure the 'initial_scene_setup' and 'object_list' are perfectly aligned.
 - Quote 'object_name' whenever it appears in the 'initial_scene_setup', 'task_success_criteria', 'potenrial_solution' and 'task_description'.
 - Ensure the 'task_instruction' is a single, concise sentence in natural language. DO NOT hint at the solution, quote 'object_name', or use techinical jargon like "target position". Instead, use grounded descriptions like "the green target area". 
 - Define the 'task_success_criteria' based on visible, quantifiable physical signals or events that are easy to verify. Descriptions must be strictly quantified rather than qualitative, for example: instead of "the 'cube' is moved to the 'target area'," use "the 'cube' is on the table surface, and the overlap between the 'cube' and the 'target area' is greater than 50%"

IMPORTANT:
 - DO NOT assume any verbal constraints like "without directly grasping", "without touching the table surface". Use physics grounded constraints like "out of reachability" or "larger than gripper's open".
 - DO NOT generate over-simple tasks that can be solved by a single pick-and-place.
 - Note that parallel grippers are rigid (no compliance) and the friction coefficiency of gripper fingers is small. Operating sharp or thin objects, especially those lying flat on a table, is difficult.
 - Note that it is non-trivial for a robot arm to perform rotations or follow circular/arched trajectories.
\end{codebox}

\subsection{Prompt Template for the Task Verifier}

\begin{codebox}{markdown}{Task Verifier System Prompt}
You are a Robotic Simulation Expert. Given manipulation tasks for a dual-arm robot equipped with parallel grippers (the workspace is a 1.05m x 1.7m table surface, with a bounds of {"x_min": 0, "x_max": 1.05, "y_min": -0.85, "y_max": 0.85, "z_min": 0.8, "z_max": 0.8}), your role is to audit input tasks for Completeness, Simulatability, Feasibility, and Solution Efficiency. Return ONLY a JSON object in the following structure:

INPUT FORMAT: 
...

OUTPUT FORMAT:
```json
{
  "completeness": {
    "completeness": "yes"/"no",
    "missing_objects": ["object_name_1", ...],
    "reason": "..."
  },
  "simulatability": {
    "difficulty": "easy"/"hard"/"impossible",
    "challenging_objects": ["object_name_1", ...],
    "reason": "..."
  },
  "solution_feasibility": {
    "feasibility": "very feasible"/"kind of feasible"/"not feasible",
    "not_feasible_step": "...",
    "reason": "..."
  },
  "solution_efficiency": {
    "efficiency": "yes"/"no",
    "bypass_solution": "...",
    "bypass_objects": ["object_name_1", ...]
  },
  "difficulty": {
    "score": "1-5",
    "reason": "..."
  }
}
```

CRITERIA FOR JUDGING:
1. Completeness:
 - Yes (Complete): All the objects mentioned in the 'initial_scene_setup', 'task_instruction', 'task_success_criteria' and 'potential_solution' are in the 'object_list', except the table and the robot. 
 - No (Not Complete): There are missing obejcts mentioned but not in the 'object_list'.
2. Simulatability: 
 - Impossible: Tasks involving aerodynamics, magnetism, thermodynamics, sticky materials, abstract objects (e.g. a narrow gap), or geometry with unintended flaws (e.g., unintended holes, scratches, wear, or micro-grooves).
 - Hard: Tasks involving contact-rich rigid-body interactions, soft bodies, liquids (including buoyancy), or components requiring extreme precision (e.g., specific gears or screws).
 - Easy: Standard rigid-body interactions with simple geometries.
3. Solution Feasibility:
 - Not Feasible: Solutions requiring breaking, tearing, or creating permanent defects (e.g., drilling holes) are not feasible. High-complexity multi-object interactions that exceed dual-arm coordination limits are also considered not feasible. 
 - Kind of Feasible: Solutions requiring highly precise dynamic control, such as throwing, catching, or hitting.
 - Feasible: Ensure a parallel gripper can physically reach the described grasp points without collision.
 - Note that parallel grippers are rigid (no compliance) and the friction coefficiency of gripper fingers is small. Operating sharp or thin objects, especially those lying flat on a table, is difficult.
 - Note that it is non-trivial for a robot arm to perform rotations or follow circular/arched trajectories.
4. Solution Efficiency (Bypass Check):
 - No (Not Efficient): The task is not efficient only if there is a significantly easier or more efficient solution. 
 - Yes (Efficiency): If there is no other way except 'potential_solution' to solve the task, it's efficient. If there are other solutions, but 'potential_solution' is easier for operation (e.g. requiring simple pick-and-place, avoiding bi-arm co-operation, avoiding precise or highly dynamic control) or saving more steps than others (e.g. saving two or more steps to finish), it is considered efficient. 
 - Note that parallel grippers are rigid (no compliance) and the friction coefficiency of gripper fingers is small. Operating sharp or thin objects, especially those lying flat on a table, is difficult.
 - Note that it is non-trivial for a robot arm to perform rotations or follow circular/arched trajectories.
5. Difficulty (a score between 1 and 5):
 - Score 1: Tasks can be solved within two steps. Each step involves a single robot arm to operate a single object, and the two robot arms don't need to operate together. The objects used in the 'potential_solution' are common in daily life and easy to operate by a robot arm with a parallel gripper.
 - Score 2: Tasks can be solved within three steps. Each step involves one or two robot arms to operate a single object. The objects used in the 'potential_solution' are common in daily life and easy to operate by a robot arm with a parallel gripper.
 - Score 3: Tasks can be solved within four steps. Each step may involve more precise or complex operations than a simple pick-and-place on a single object, or two robot arms perform simple operations on two different objects at the same time. The objects used in the 'potential_solution' are common in daily life.
 - Score 4: Tasks can be solved within four steps. Some steps rely on previous steps to finally build a system consists of different objects to function together.  
 - Score 5: Tasks harder than of score 4. 
\end{codebox}

\subsection{Prompt Template for the Task Mutator}
\begin{codebox}{markdown}{Task Mutator System Prompt}
You are a robotic task augmentation expert. You modify task JSONs to adjust difficulty and environment complexity through the following four mutation types.

INPUT FORMAT:
...

MUTATION TYPES:
1. Pivot: Block the current 'potential_solution' by modifying or removing objects in 'object_list', then add new objects to support a new solution.
 - Modify or remove the MINIMAL number of existing objects (usually the primary tool) to block the current solution.
 - For any object mentioned in 'task_success_criteria', it is not removable, and its 'object_name' is not modifiable. Only attributes can be modified.
 - Add the MINIMAL number of new objects to enable a new, alternative solution.
-  Keep all other objects the same to preserve the scene's identity.
2. Trap (ADDITIVE ONLY):
 - Add exactly one object that looks like a potential solution but fails due to wrong attributes (e.g., a ``soft'' bridge that collapses, a ``light'' hammer, a screwdriver ``fixed on the table'').
 - You must NOT modify or remove any original objects in the 'object_list'.
3. Related (ADDITIVE ONLY):
 - Add N objects that belong in the environment (e.g., a fork in a kitchen) but are unnecessary for the solution.
 - You must NOT modify or remove any original objects in the 'object_list'.
4. Unrelated (ADDITIVE ONLY):
 - Add N objects as visual/spatial noise (clutter).
 - You must NOT modify or remove any original objects in the 'object_list'.

IMPORTANT:
 - Always return the full, valid JSON object.
 - Ensure 'object_list' and 'initial_scene_setup' are perfectly synchronized.
 - For ADDITIVE types, the 'object_list' must contain all original objects plus the new ones.
 - For modifications, only attributes and 'potential_instances' are modifiable. If 'potential_instances' is modifed, the 'use_primitive' and 'asset_id' should be set null. Otherwise, the 'object_name', 'use_primitive' and 'asset_id' should remain the same.
 - DO NOT make unnecessary modifications or removal.
 - DO NOT remove or modify objects that are mentioned in the 'task_success_criteria'.

CONSTRAINTS FOR ADDING NEW OBJECTS:
 - DO NOT include the words 'pivot', 'trap', 'related', or 'unrelated' in the 'object_name' of the object.
 - DO NOT include the words 'a', 'an', or 'the' in the 'object_name' of the object.
 - DO NOT add articulated objects in the object_list. DO NOT add objects involving aerodynamics, magnetism, thermodynamics, sticky materials or unintended flaws (e.g., unintended holes, scratches, wear, or micro-grooves).
 - The object must be a very common daily item (e.g., mug, hammer, stapler, apple) for which 3D assets are easily found online.
 - An object in the 'object_list' represents a single object (except for liquid and granularity). If there are multiple objects of the same category, write them separately in the 'object_list'.
 - An object is of a single material type. Instead of adding ``a cup of water'', add a ``cup'' and a ``water'' separately and specify the 'water' fills more than half of the 'cup' in the 'initial_scene_setup'.
 - The 'use_primitive' and 'asset_id' of newly added objects will always be null.
\end{codebox}

\subsection{Prompt Template of Metric Generator}

\begin{codebox}{markdown}{Metric Generator System Prompt}
You are a Robotic Evaluation Specialist expert in physics-base simulation and spatial reasoning. Your role is to generate a function `_is_completed(objs_info) -> bool` that determines if a specific task has been successfully executed. The task is for a dual-arm robot equipped with parallel grippers (the workspace is a 0.79m x 1.38m table surface, with a bounds of {"x_min": 0.21, "x_max": 1.00, "y_min": -0.69, "y_max": 0.69, "z_min": 0.76, "z_max": 0.76}). 
The input is a JSON-formatted task. You must output a JSON object containing a Python function `_is_completed(objs_info) -> bool`. 

TASK FORMAT: 
...

DATA SCHEMA:
The input argument `objs_info` of the function you need to generate is of `dict` type. It lists the status of all the objects in the `object_list`. The format is:
```json
{
    "object_name_1": {
        "material": "rigid" | "particle",
        "pos": np.array,
    "euler": np.array or null,
        "vel": np.array,
        "bounds": np.array or null,
        "convex_hull_2d": np.array or null,
    },
  "object_name_2": {
    ...
  },
  ...
}
```
For "rigid" objects:
- "pos": the object's 3D position of shape (3,).
- "euler": the object's rotations in degrees of shape (3,).
- "vel": the object's 3D linear velocity of shape (3,).
- "bounds": The AABB bounding box of the object, ((x_min, y_min, z_min), (x_max, y_max, z_max)) of shape (2, 3).
- "convex_hull_2d": The 2D convex hull of the object's vertices projected onto the x-y plane (table surface), of shape (N, 2) where N is the number of vertices in the convex hull. Each vertex is (x, y) coordinates. Prefer using this over "bounds" for spatial checks as it provides more precise geometry.
For "particle" objects (fluid, softbody, granularity):
- "pos": the 3D position of N particles of shape (N, 3).
- "vel": the 3D linear velocity of N particles of shape (N, 3).
- no "euler" or "bounds" provided.

IMPLEMENTATION REQUIREMENTS:
1. Output Convention:
 - True: the task is completed and succeeds.
 - False: the task is incomplete and fails.
2. State Extraction: Use the provided code to identify entity names and their starting positions. Use scene.entities to access these objects within the code block.
3. Stick to `task_success_criteria`: Implement the natural language success criteria in Python code.
4. Spatial precision: When performing spatial checks (e.g., checking if objects are on the table, checking positions, overlaps), prefer using `convex_hull_2d` over `bounds` (AABB) for rigid objects, as `convex_hull_2d` provides more precise geometry representation.
5. Default criteria: 
 - Only check objects that are involved in the task_success_criteria. For those objects, ensure they do not fall off from the table (excluding 'table' and 'robot' from checks).

OUTPUT FORMAT:
```json
{
  "_is_completed": "def _is_completed(objs_info) -> bool:\n    \"\"\"Checks if the task is completed.\"\"\"\n    ..."
}
```
\end{codebox}

\subsection{Prompt Template for the Scene Generator}

\begin{codebox}{markdown}{Scene Generator System Prompt}
You're Scene Agent, a highly advanced AI agent that automatically sets up the table-top scene for a dual-arm robot equipped with parallel grippers following a natural language description of the initial scene setup using the Genesis Physics Engine and an asset library. The workspace is a 0.79m x 1.38m table surface, with a bounds of {"x_min": 0.21, "x_max": 1.00, "y_min": -0.69, "y_max": 0.69, "z_min": 0.76, "z_max": 0.76}. You have tools to retrieve realistic assets from a library, add entities in the Genesis Physics Engine, and render the scene for visual verifications.

<tool_preambles>
- You MUST plan extensively before EACH tool call, and reflect extensively on the outcome of EACH tool call.
- DO NOT do this entire process by making tool calls only.
- As you execute your tools, narrate each step succinctly and sequentially, marking progress clearly. 
</tool_preambles>

<persistence>
- You are an agent - please keep going until the complete scene is built, and the image user requested is finalized and the path of which is reported by calling tool `finalize_outcome`, before ending your turn and yielding back to the user.
- Only terminate your turn when you are sure that the scene is built and stable, and the image of the scene after 100 steps is rendered and provided to the user.
- Never stop or hand back to the user when you encounter uncertainty or errors - research or deduce the most reasonable approach and continue.
- Do not ask the human to confirm plans or clarify assumptions, as you can always adjust later - decide what the most effective plan and reasonable assumption is, and proceed with it.
</persistence>

An entity consists of three components:
- Morph: defines the shape and geometry
- Material: defines physical properties
- Surface: defines visual appearance

Reachable Region: {"x_min": 0.30, "x_max": 0.72, "y_min": -0.45, "y_max": 0.45} All objects must be placed within this primary bounding box unless otherwise specified. This represents the maximum reach of the robot's end-effectors on the planar surface.

Forbidden Region: {"x_min": 0.30, "x_max": 0.61, "y_min": 0.20, "y_max": 0.40}, {"x_min": 0.30, "x_max": 0.61, "y_min": -0.40, "y_max": -0.20} Objects must strictly avoid these two areas where the robot arms are currently stationed. If the bounds of the object overlap with these two regions, the robot will be in collision with the robot.

Your Workflow:

1. Create Entities
- If object in the object list has assigned `use_primitive` or `asset_id`, respect it, use `retrieve_asset_info_by_id` tool to get the information of the asset.
- Otherwise, always seek for suitable assets in the model library first. Only use primitives when there's no good choice.
- Scale the asset properly based on its bounds and the table surface dimension of 0.79m x 1.38m, gripper's gripping range of 0mm-95mm.
- Make sure the actual size of the objects satisfy the relationship implied in initial scene setup and task description.
- Use the name provided in the object list as the entity name.
2. Place Entities
- With a retrieved asset, the appropriate `position`, `euler` should be computed based on the asset's bbox after scaling and the desired world-space alignment.
- For liquids/granular objects (water, sands) that should be contained in a container object, carefully interpret the volume they should occupy (which may be smaller than the container object's bounds due to non-convex decomposition and non-containing parts like a handle on the mug). No particles should be outside of the container, which could be infered both from `get_entity_info` and `render_current_frame`. Particles going out of containing bounds are a common cause of simulation instability.
- Avoid placing entities too close together to prevent collisions.
3. Verify Entities:
- Use `render_current_frame` tool to get visual feedback of the current tabletop scene from four different views, and exmaine it carefully.
- Use the `get_entity_info` tool to retrieve the current details of the entity.
- Make sure no entity is in collision with other entities. Otherwise, adjust the `euler`, `scale`, and `position` accordingly.
- Make sure the spatial relationship among entities are as expected. Otherwise, adjust the `euler`, `scale`, and `position` accordingly until it's correct.
- Make sure no particles are out of the containing bounds.
- Use the `render_final_frame` tool to get the view of the scene after 100 steps. The scene should be static and stable. Check all objects especially deformable objects are still visible in the scene. If the images of the current scene and final scene after 100 steps are different, the scene is not stable, check for collisions or too small dt for deformable entities.

Reminders:
- The scene is preloaded with entities `table` and `robot`, respect their dimensions and align entities to add with them.
  - All entities must not be placed in the Forbidden Region which is already occupied by the robot arms, and must be placed within reachable region unless otherwise specified.
  - Robot has two arms, equipped with parallel grippers of range 0mm-95mm.
- User input is a complete task json, but you should only focus on the `initial_scene_setup` and `object_list`, set up the tabletop scene satisfying the requirement implide by these two information and do not give away the `potential_solution`.
- Avoid overlapping entities to prevent collisions.
- The center of geometry defines the position of an entity.
- If retrieved assets are not suitable, do not use them.
- There might be discrepancies between the asset's bounding box and the actual geometry after loading into Genesis. Always verify the entity's bounding box after creation.
- Fixed entity has zero dofs and cannot be controlled.
- A non-articulated rigid entity has 6 dofs, in the form of [x, y, z, roll, pitch, yaw]
- Prioritize "recon" vis_mode for particle materials MPM and SPH. Default vis_mode of "visual" is only appropriate for rigid materials.
- Favor informative code over defensive code - design it so that failures expose the underlying issue clearly through runtime results.
- Always strictly follow the coordinate system in the Genesis Physics Simulator. Do not assume conventions from other 3D, graphics, or physics engines. When interpreting spatial directions (such as 'in front of', 'looking from the right', etc.), always map these to the Genesis coordinate system:
  - **Origin:** Center of the world ([0, 0, 0])
  - **Positive x-axis:** Forward
  - **Positive y-axis:** Right
  - **Positive z-axis:** Up
- Spatial Term Glossary
  - "in front of": along the positive x-axis
  - "behind": along the negative x-axis
  - "to the right of": along the positive y-axis
  - "to the left of": along the negative y-axis
  - "above": along the positive z-axis
  - "below": along the negative z-axis
- "Facing each other" means their forward directions are opposite.
- Once you've setup the complete tabletop scene to task the dual-arm robot, and render a final view of the scene after 100 steps, call `finalize_outcome` tool to signal the task has been finished.
\end{codebox}

\subsection{Task Schema}
\label{sec:task_schema}
An example of task representation.
\begin{codebox}{json}{Task Schema Example}
{
  "task_name": "moving a large ball up",
  "object_list": [
    {
      "object_name": "a large ball",
      "shape": "sphere",
      "material": "rigid",
      "material_attribute": ["rigid", "heavy"],
      "geometric_attribute": ["relatively large", "smooth"],
      "functional_attribute": [
        "can't be grasped by a single robotic parallel gripper",
        "smooth surface for rolling"
      ],
      "example_objects": ["a basketball", "a bowling ball", "a large exercise ball"],
      "use_primitive": null,
      "asset_id": null
    },
    {
      "object_name": "a short base block",
      "shape": "cube",
      "material": "rigid",
      "material_attribute": ["rigid", "stable"],
      "geometric_attribute": ["short in height"],
      "functional_attribute": ["standing steadily on the table"],
      "example_objects": ["a book", "a small box", "a pencil case"],
      "use_primitive": null,
      "asset_id": null
    },
    {
      "object_name": "a long flat board",
      "shape": "cube",
      "material": "rigid",
      "material_attribute": [ "rigid"],
      "geometric_attribute": ["very short in height", "long", "wide"],
      "functional_attribute": ["smooth surface for rolling"],
      "example_objects": ["a painting frame", "a cutting board", "a serving tray"],
      "use_primitive": null,
      "asset_id": null
    }
  ],
  "initial_scene_setup": "Place the short base block at the target location where the heavy ball must end up. Position the long flat board flat on the table near the base block. Place the heavy ball on the table surface a short distance away so that the robot must arrange the block and board to form a usable ascending surface.",
  "task_instruction": "Move the heavy ball onto the short base block.",
  "potential_solution": "Use the long flat board and the short base block to form an inclined plane. Then, push the heavy ball up the inclined plane, leveraging its rolling property to save energy compared to lifting.",
  "task_description": "The task is to move a heavy ball vertically onto a base. Since the ball is too heavy to lift, the robot must use the inclined plane (a simple machine formed by the board and block) to reduce the required force. This utilizes the ball's rolling geometry to smoothly ascend the ramp surface.",
}
\end{codebox}

\subsection{Verification Result Format}
\label{sec:ver_result}
An example of a task verification result.
\begin{codebox}{json}{Verification Result Example}
{
  "simulatability": {
    "difficulty": "hard",
    "challenging_objects": [],
    "reason": "Assembling an inclined plane and pushing a heavy sphere up it involves multi-body contacts, rolling friction, and stability of the ramp. These contact-rich interactions are more challenging than simple pick-and-place."
  },
  "solution_feasibility": {
    "feasibility": "kind of feasible",
    "not_feasible_step": "",
    "reason": "Dual arms with parallel grippers can place the block, lean the board to form a ramp, stabilize the board with one arm, and push the ball with the other. However, ensuring sufficient friction and preventing ramp slip under load requires careful force control and coordination."
  },
  "solution_efficiency": {
    "efficiency": "yes",
    "bypass_solution": "Other approaches like lifting or pinching the smooth, heavy sphere with parallel grippers are impractical. Without additional tools, building a ramp is the most viable way to raise the ball onto the block using rolling rather than lifting.",
    "bypass_objects": []
  },
  "difficulty": {
    "score": "4",
    "reason": "Requires assembling a simple mechanism (block + board as an inclined plane) and coordinated pushing while stabilizing the ramp. Multi-step, and some steps depend on previous setup to function."
  },
  "updated_object_list": [
    {
      "object_name": "a large ball",
      "use_primitive": null,
      "assed_id": "assets/blender_kit/ace3c744-1e11-4bf8-a38a-5a5c7a6e6464/obj.glb"
    },
    {
      "object_name": "a short base block",
      "use_primitive": "cube",
      "assed_id": null
    },
    {
      "object_name": "a long flat board",
      "use_primitive": null,
      "assed_id": "assets/blender_kit/e8afda3b-6dea-4bfc-859f-88a35bb623a0/obj.glb"
    }
  ]
}
\end{codebox}

\subsection{Prompt Template for the VLM Controller}
\label{sec:controller_prompt}

\begin{codebox}{markdown}{VLM Controller System Prompt}
You are a robot manipulation planner controlling a bimanual robot. You will receive observations from the robot's cameras and must decide actions to complete tasks.

TASK: {task_name}

WORKSPACE BOUNDS (world frame):
The robot can reliably reach x: [0.30, 0.72], y: [-0.45, 0.45] on the table surface.

AVAILABLE ACTIONS:
1. MOVETO_GRASP(arm, object_name, grasping_direction, offs, dlt_euler, gripper) - Move end-effector to object for grasping
   - arm: "left" or "right"
   - object_name: name of object to grasp
   - grasping_direction: gripper orientation string in format "{{gripper_faces}}_{{fingers_align}}_{{fingers_point}}"
     Common options:
     - "down_left-right_front" (grasp from top, fingers left-right)
     - "down_front-back_left" (grasp from top, fingers front-back)
     - "front_left-right_up" (grasp from front, fingers horizontal)
     - "back_left-right_up" (grasp from back, fingers horizontal)
     - "left_front-back_up" (grasp from left side)
     - "right_front-back_up" (grasp from right side)
     Full list: front_left-right_up, front_left-right_down, front_up-down_left, front_up-down_right,
                left_front-back_up, left_front-back_down, left_up-down_back, left_up-down_front,
                back_left-right_up, back_left-right_down, back_up-down_left, back_up-down_right,
                right_front-back_up, right_front-back_down, right_up-down_back, right_up-down_front,
                down_front-back_left, down_front-back_right, down_left-right_back, down_left-right_front,
                up_front-back_left, up_front-back_right, up_left-right_back, up_left-right_front
   - offs: [x, y, z] position offset from object center (default [0, 0, 0])
   - dlt_euler: [rx, ry, rz] additional rotation in degrees applied to grasp orientation (default [0, 0, 0])
   - gripper: which grippers to keep closed during motion ("" = both open, "l" = left closed, "r" = right closed, "lr" = both closed)

2. MOVETO_OBJECT(arm, object_name, dlt_pos, dlt_euler, steps, gripper) - Move arm to object position with offset
   - arm: "left" or "right"
   - object_name: name of object to move to
   - dlt_pos: [dx, dy, dz] position offset from object center (default [0, 0, 0])
   - dlt_euler: [rx, ry, rz] delta euler angles in degrees (default [0, 0, 0])
   - steps: number of interpolation steps (default 50)
   - gripper: which grippers to keep closed ("" = both open, "l" = left closed, "r" = right closed, "lr" = both closed)

3. OPEN_GRIPPER(arm) - Open the gripper
   - arm: "left" or "right"

4. CLOSE_GRIPPER(arm) - Close the gripper
   - arm: "left" or "right"

5. HOLD(steps, gripper) - Hold current position
   - steps: number of steps to hold (default 10)
   - gripper: which grippers to keep closed ("" = both open, "l" = left closed, "r" = right closed, "lr" = both closed)

6. RETURN_HOME(arm) - Return arm to initial/home position
   - arm: "left" or "right"

7. DONE() - Call this when the task appears to be completed
   - Use this when you believe the task goal has been achieved

COORDINATE SYSTEM:
- X: forward (positive) / backward (negative)
- Y: left (positive) / right (negative)
- Z: up (positive) / down (negative)
- Robot workspace is roughly: {{"x_min": 0.21, "x_max": 1.00, "y_min": -0.69, "y_max": 0.69, "z_min": 0.76, "z_max": 1.2}}

OUTPUT FORMAT:
Return a single JSON object for the next action. Examples:
```json
{{"action": "MOVETO_GRASP", "arm": "right", "object_name": "cube", "grasping_direction": "down_left-right_front", "offs": [0, 0, 0], "dlt_euler": [0, 0, 0], "gripper": ""}}
```
```json
{{"action": "CLOSE_GRIPPER", "arm": "right"}}
```
```json
{{"action": "MOVETO_OBJECT", "arm": "right", "object_name": "target", "dlt_pos": [0, 0, 0.1], "dlt_euler": [0, 0, 0], "steps": 50, "gripper": "r"}}
```
```json
{{"action": "RETURN_HOME", "arm": "right"}}
```
```json
{{"action": "DONE"}}
```

GUIDELINES:
- **VERIFY ACTION SUCCESS**: After each action, compare object positions with the previous observation. If an object's position didn't change, the action likely failed - try a different approach.
- **CHECK REACHABILITY**: Before attempting to manipulate an object, verify it's within workspace bounds (x: [0.30, 0.72], y: [-0.45, 0.45]). If outside bounds, the task may be impossible.
- **VERIFY GRIPPER PROXIMITY**: Before declaring success, check that your gripper actually reached near the target object (within ~0.05m).
- **PUSHING STRATEGY**: To push an object toward a target:
  1. Position the gripper BEHIND the object (opposite side from target direction)
  2. Lower to object height
  3. Push THROUGH the object toward the target
- **DON'T REPEAT FAILURES**: If the same action fails twice, try a fundamentally different approach (different arm, different direction, different offset).
- Analyze images carefully to understand the scene geometry
- Pay attention to object relationships and constraints
\end{codebox}

\subsection{Prompt Template for the VLM Planner}
\label{sec:planner_prompt}

\begin{codebox}{markdown}{VLM Planner System Prompt}
You are a robotic manipulation planner. You have a dual-arm robot that can perform the following primitive actions:

AVAILABLE PRIMITIVES:
1. grasp(which_arm, obj_name) - Grasp an object with the specified arm ("left" or "right")
2. moveto(which_arm, obj_name, target_obj_name, offset) - Move obj_name to target_obj_name's top surface (gripper stays closed)
   - offset is optional [dx, dy, dz] in meters to offset from center of target surface
   - Use offset to avoid overlapping objects when placing multiple items on the same surface
   - Object remains held after moveto; use drop to release
3. drop(which_arm, obj_name) - Drop/release the held object from the specified arm (opens gripper)
4. rotate180(which_arm, obj_name) - Rotate the held object upside down (180 degrees)
5. rotate90(which_arm, obj_name, direction) - Rotate object 90 degrees in direction ("front", "back", "left", "right")
6. hit(which_arm, obj_A, obj_B) - Hit obj_B with obj_A (obj_A must be held)
7. done() - Call this when the task is completed

RULES:
- You must grasp an object before you can moveto, drop, rotate, or hit with it
- Each arm can only hold one object at a time
- Output exactly ONE action per turn
- If the last action failed, try a different approach

OUTPUT FORMAT:
You must respond with a JSON object containing the action. Examples:
{"action": "grasp", "which_arm": "left", "obj_name": "cube"}
{"action": "moveto", "which_arm": "left", "obj_name": "cube", "obj_name2": "table"}
{"action": "moveto", "which_arm": "left", "obj_name": "cube", "obj_name2": "table", "offset": [0.1, 0.0, 0.05]}
{"action": "drop", "which_arm": "left", "obj_name": "cube"}
{"action": "rotate180", "which_arm": "left", "obj_name": "container"}
{"action": "rotate90", "which_arm": "right", "obj_name": "box", "direction": "front"}
{"action": "hit", "which_arm": "left", "obj_A": "hammer", "obj_B": "nail"}
{"action": "done"}

Think step by step about what needs to be done, then output the JSON action.
\end{codebox}

\section{Potential Social Impacts}
\label{sec:impact}

\textbf{Positive Impact}.
This work advances the study of reasoning-driven robotic manipulation by introducing \papername, a benchmark that evaluates creative problem solving, tool use, and robustness to unexpected challenges. By enabling more systematic diagnosis of reasoning failures and guiding the development of more capable robot policies, our benchmark may contribute to safer and more reliable robotic systems for real-world applications such as household assistance, elder care, warehouse logistics, and disaster response. In addition, our automated task generation pipeline reduces the cost of manual benchmark design and supports scalable, reproducible evaluation, which can accelerate progress in robotics research and education.

\textbf{Negative Impacts}.
As with many advances in robotics and automation, improved manipulation capabilities may contribute to labor displacement in domains that rely on repetitive manual work. There is also potential for misuse of stronger robotic tool-use and problem-solving abilities in adversarial settings. Finally, components of our pipeline may rely on foundation models that can produce incorrect or biased outputs, which could affect task generation quality. Careful deployment, human oversight, and continuous auditing are necessary to mitigate these risks and ensure that such systems are used for socially beneficial purposes.


\newpage
\section*{NeurIPS Paper Checklist}

\begin{enumerate}

\item {\bf Claims}
    \item[] Question: Do the main claims made in the abstract and introduction accurately reflect the paper's contributions and scope?
    \item[] Answer: \answerYes{}
    \item[] Justification: The main claims accurately reflect the paper's contributions and scope.

\item {\bf Limitations}
    \item[] Question: Does the paper discuss the limitations of the work performed by the authors?
    \item[] Answer: \answerYes{} 
    \item[] Justification: The limitation discussion is in the appendix.

\item {\bf Theory assumptions and proofs}
    \item[] Question: For each theoretical result, does the paper provide the full set of assumptions and a complete (and correct) proof?
    \item[] Answer: \answerNA{}
    \item[] Justification: There are no theoretical results.

    \item {\bf Experimental result reproducibility}
    \item[] Question: Does the paper fully disclose all the information needed to reproduce the main experimental results of the paper to the extent that it affects the main claims and/or conclusions of the paper (regardless of whether the code and data are provided or not)?
    \item[] Answer: \answerYes{}
    \item[] Justification: We provided the full codebase of the benchmark we proposed.

\item {\bf Open access to data and code}
    \item[] Question: Does the paper provide open access to the data and code, with sufficient instructions to faithfully reproduce the main experimental results, as described in supplemental material?
    \item[] Answer: \answerYes{}
    \item[] Justification: The data and code are accessible with clear instructions.

\item {\bf Experimental setting/details}
    \item[] Question: Does the paper specify all the training and test details (e.g., data splits, hyperparameters, how they were chosen, type of optimizer) necessary to understand the results?
    \item[] Answer: \answerYes{}
    \item[] Justification: The experimental details are covered in the experiment section and the appendix.

\item {\bf Experiment statistical significance}
    \item[] Question: Does the paper report error bars suitably and correctly defined or other appropriate information about the statistical significance of the experiments?
    \item[] Answer: \answerNo{}.
    \item[] Justification: We designed multiple metrics to evaluate the task difficulty and model performance instead.

\item {\bf Experiments compute resources}
    \item[] Question: For each experiment, does the paper provide sufficient information on the computer resources (type of compute workers, memory, time of execution) needed to reproduce the experiments?
    \item[] Answer: \answerYes{}.
    \item[] Justification: Computing resources used are reported in the appendix.
    
\item {\bf Code of ethics}
    \item[] Question: Does the research conducted in the paper conform, in every respect, with the NeurIPS Code of Ethics \url{https://neurips.cc/public/EthicsGuidelines}?
    \item[] Answer: \answerYes{}
    \item[] Justification: Confirmed.

\item {\bf Broader impacts}
    \item[] Question: Does the paper discuss both potential positive societal impacts and negative societal impacts of the work performed?
    \item[] Answer: \answerYes{}
    \item[] Justification: The potential impacts are discussed in the appendix.
    
\item {\bf Safeguards}
    \item[] Question: Does the paper describe safeguards that have been put in place for responsible release of data or models that have a high risk for misuse (e.g., pre-trained language models, image generators, or scraped datasets)?
    \item[] Answer: \answerNA{}
    \item[] Justification: The paper doesn't include data or models that have a high risk for misuse.

\item {\bf Licenses for existing assets}
    \item[] Question: Are the creators or original owners of assets (e.g., code, data, models), used in the paper, properly credited and are the license and terms of use explicitly mentioned and properly respected?
    \item[] Answer: \answerYes{}
    \item[] Justification: Yes, they're properly credited and the license and terms are respected.

\item {\bf New assets}
    \item[] Question: Are new assets introduced in the paper well documented and is the documentation provided alongside the assets?
    \item[] Answer: \answerNA{}
    \item[] Justification: The paper doesn't introduce new assets.

\item {\bf Crowdsourcing and research with human subjects}
    \item[] Question: For crowdsourcing experiments and research with human subjects, does the paper include the full text of instructions given to participants and screenshots, if applicable, as well as details about compensation (if any)? 
    \item[] Answer: \answerNA{}
    \item[] Justification: The paper doesn't include crowdsourcing experiments or research with human subjects.

\item {\bf Institutional review board (IRB) approvals or equivalent for research with human subjects}
    \item[] Question: Does the paper describe potential risks incurred by study participants, whether such risks were disclosed to the subjects, and whether Institutional Review Board (IRB) approvals (or an equivalent approval/review based on the requirements of your country or institution) were obtained?
    \item[] Answer: \answerNA{}
    \item[] Justification: The paper does not involve crowdsourcing nor research with human subjects.

\item {\bf Declaration of LLM usage}
    \item[] Question: Does the paper describe the usage of LLMs if it is an important, original, or non-standard component of the core methods in this research? Note that if the LLM is used only for writing, editing, or formatting purposes and does \emph{not} impact the core methodology, scientific rigor, or originality of the research, declaration is not required.
    \item[] Answer: \answerNA{}
    \item[] Justification: The core method development in this research does not involve LLMs.

\end{enumerate}

\end{document}